\documentclass{article} 
\usepackage{iclr2027_conference,times}

\usepackage{amsmath,amsfonts,bm}

\def\eqref#1{equation~\ref{#1}}

\def\1{\bm{1}}

\DeclareMathAlphabet{\mathsfit}{\encodingdefault}{\sfdefault}{m}{sl}
\SetMathAlphabet{\mathsfit}{bold}{\encodingdefault}{\sfdefault}{bx}{n}

\usepackage{hyperref}
\usepackage{url}

\usepackage{graphicx}
\usepackage{booktabs}
\usepackage{amssymb}
\usepackage{wrapfig}
\usepackage{multirow}
\usepackage{subcaption}
\usepackage{xcolor}

\title{Generative Embodied Multiple Behavior Control Systems for Human-like Agents}

\author{
\textbf{Chongyu Bao}\textsuperscript{1,*},
\textbf{Haokai Yang}\textsuperscript{1,*},
\textbf{Yuhan Wang}\textsuperscript{1},
\textbf{Zhaochong An}\textsuperscript{2},
\textbf{Kunpeng Liu}\textsuperscript{3},
\textbf{Xiaolan Liu}\textsuperscript{1}
\\[4pt]
\normalfont\textsuperscript{1}Bristol Digital Futures Institute, University of Bristol
\\
\normalfont\textsuperscript{2}University of Copenhagen
\\
\normalfont\textsuperscript{3}Clemson University
\\[4pt]
\normalfont\textsuperscript{*}Equal contribution
}

\iclrfinalcopy

\begin{document}

\maketitle

\vspace{-1em}
\begin{center}
    \includegraphics[width=\textwidth]{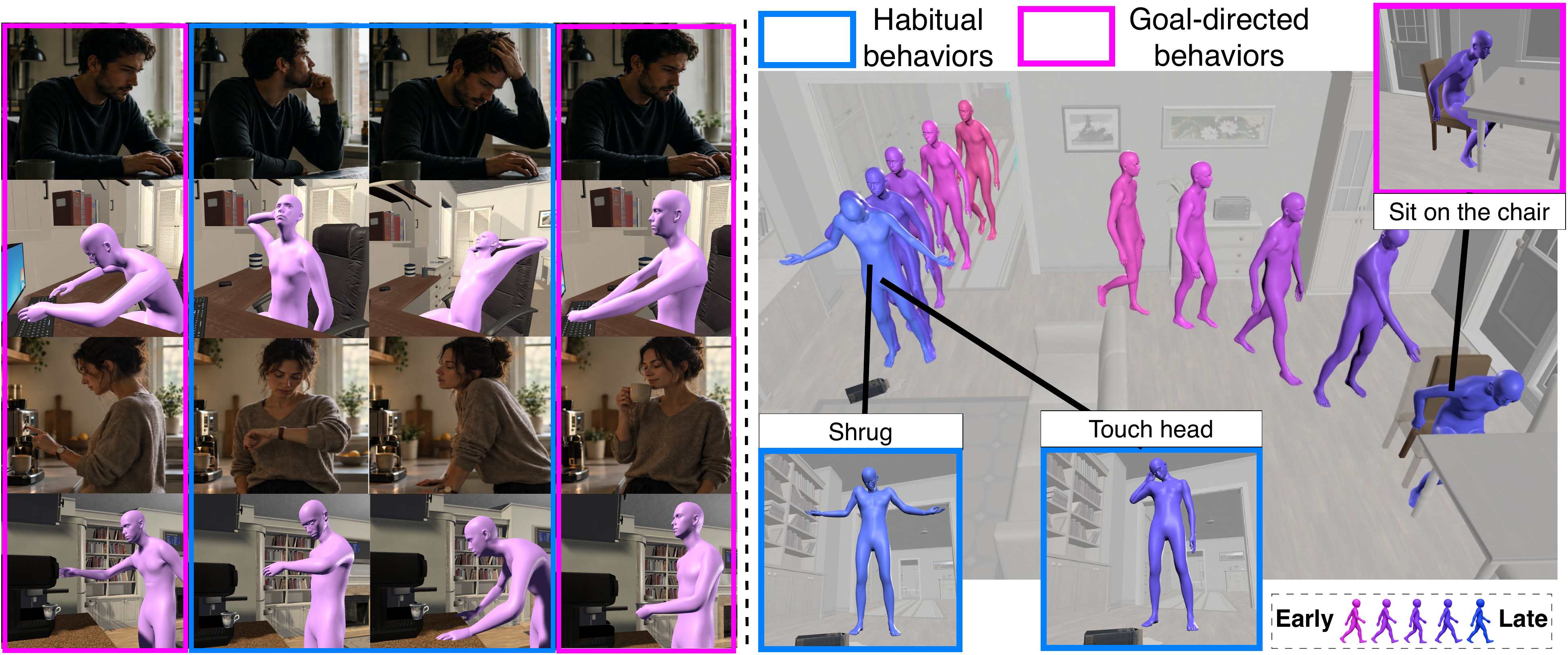}
    \captionof{figure}{\textbf{Believable human-like behaviors with GEMS.}
        \textit{(Left.)} Comparison between real human daily behaviors and GEMS-generated behaviors. Beyond goal-directed behaviors, GEMS generates habitual behaviors commonly observed in everyday human life but often overlooked. \textit{(Right.)} Examples of motion sequences generated by GEMS. Our method generates diverse habitual and goal-directed behaviors and realizes them as coherent human–scene interactions in 3D environments.
    }
    \label{fig:cognitive_architecture}
\end{center}
\vspace{-0.5em}

\begin{abstract} 
Building autonomous agents that reproduce human behavior in realistic 3D environments has been a longstanding objective in AI. Cognitive neuroscience commonly believes that human behaviors are controlled by multiple control systems including goal-directed and habitual. However, existing human-like agent frameworks primarily focus on goal-directed system and habitual system is largely overlooked.
In this paper, we address this gap by proposing \textbf{GEMS}, a \textbf{G}enerative \textbf{E}mbodied \textbf{M}ultiple Behavior Control \textbf{S}ystem that jointly models goal-directed and habitual behaviors for generating believable human-like behavior. GEMS employs a cognitive module to generate natural language behaviors and a motion generation module to construct them in 3D environment.
The cognitive module comprises a Habitual Controller that retrieves habitual actions from habit memory in response to environmental stimuli, a Goal-directed Controller that proposes goal-directed actions for current goal, and an Arbiter that governs their influence and selects the final action, supporting diverse and naturalistic behavior patterns generation. To construct these behaviors, the motion generation module adopts a Text-to-Motion model to generate versatile and expressive 3D human–scene interactions without relying on predefined actions. Extensive evaluations, human and ablation studies demonstrate that human-likeness performance is substantially improved by GEMS. The efficacy of GEMS indicates the benefits of
leveraging habitual behavior and multiple behavior control system coordination for believable embodied human-like agents design.
The code is available at
\href{https://kilo377.github.io/GEMS/?v=20261016-clean17\#interactive-demo}
{\underline{this link}}.

\end{abstract}

\section{Introduction}

Building human-like autonomous agents \citep{sloman1999sort, sloman2000architectural} that can reproduce human behavior in realistic 3D environments has been a long-standing pursuit since the inception of artificial intelligence (AI) \citep{fetzer1990artificial}. The study could empower non-player game characters \citep{li2025x}, underpin human-robot interaction and cooperation, populate virtual reality communities \citep{zhou2026virtual}, and accelerate Embodied AI \citep{feng2026multi}.
With recent advances in large language models (LLMs), building embodied agents that can simulate believable \citep{bates1994role} and coherent human-like behavior in daily life has attracted increasing attention.
%
Existing frameworks  \citep{park2023generative,wang2024d2a,leng2026agentsense,liang2025actor,zhang2025ella,lin2026asvo,li2025x} typically focus on modeling goal-directed behaviors \citep{wood2022habits}, which are generated by considering the outcomes that actions are likely to bring about, such as satisfying desires \citep{wang2024d2a} or completing plans \citep{park2023generative}. 

However, these methods largely overlook habitual behavior, which accounts for a substantial proportion of everyday human behavior \citep{rebar2025habitual,wood2016psychology}.
Insights from cognitive psychology and neuroscience \citep{miller2019habits,dolan2013goals,lee2014neural} suggest that human behavior is governed by multiple control systems, including goal-directed and habitual control systems. \citep{rebar2025habitual,wood2016psychology}. Specifically, these systems generate behaviors through fundamentally different mechanisms. The goal-directed system generates actions that are expected to produce desired outcomes, whereas the habitual system triggers actions in response to relevant environmental stimuli.
Multiple control systems jointly shape human behavior, contributing to the diverse and natural patterns observed in everyday human activity.
Motivated by these observations, a pertinent question arises: \textit{How can we integrate multiple behavioral control systems to generate believable human-like behaviors when building human-like agents?}
Addressing this question poses three key challenges: 
(1) how to model habitual behavior;
(2) how to exploit multiple control systems jointly shape the final human behavior; and
(3) how to construct diverse habitual and goal-directed behaviors naturally and accurately in 3D environments.

In this work, we propose a \textbf{G}enerative \textbf{E}mbodied \textbf{M}ultiple Behavior Control \textbf{S}ystem (\textbf{GEMS}) for building human-like agent, which jointly models goal-directed and habitual behaviors, enabling believable 3D human-like behavior generation. GEMS employs a cognitive module to generate natural language behaviors and a motion generation module to construct them in 3D environment.
The cognitive module reflects the multiple control mechanisms underlying human behavior through three interacting components.
The \textbf{Habitual Controller} models the stimulus--response nature of habitual behavior by retrieving habitual actions associated with relevant environmental cues from a habit memory synthesized from prior empirical studies.
In parallel, the \textbf{Goal-directed Controller} uses an internal world model \citep{wu2023pre} to predict action consequences, supporting goal-directed reasoning in open-ended embodied environments. The \textbf{Arbiter} dynamically governs the relative influence of each controller and determines the final action. Together, these components integrate behaviors driven by different control mechanisms in a plausible manner, thereby enabling diverse and naturalistic behavior generation.

In addition, the motion generation module performs 3D human–scene interaction generation based on the resulting natural-language behaviors. Existing language-conditioned interaction methods \citep{wang2022humanise, wu2025human,zou2026infbagel} have limited ability to cover the diverse behaviors produced by our cognitive module. We therefore adopt a scene-agnostic Text-to-Motion (T2M) model \citep{rempe2026kimodo} for versatile and expressive motion synthesis. Since such models lack explicit scene awareness, we further introduce keyframe-guided conditioning to impose spatial and interaction constraints from the environment. Compared with existing human-like agents that rely on predefined action spaces, GEMS supports more flexible and varied behavior generation.
%

We systematically compare our method against existing human-like agent frameworks \citep{leng2026agentsense,lin2026asvo,liang2025actor,wang2024d2a,yu2024affordable} through comprehensive quantitative evaluations and human studies with \textbf{94 participants}, demonstrating its superiority in generating believable human-like behaviors.
With extensive ablation studies (\S \ref{ablation}), we offer insights into the efficacy of our designs and showcase the benefits of multiple behavior control systems.

\textbf{Our contributions are three-fold.}
(i) We introduce a real human-inspired multiple behavioral control system for believable human-like daily behavior simulation by jointly modeling goal-directed and habitual behaviors.
(ii) We propose GEMS, a human-like agent framework to effectively generate human-like behaviors in 3D environments, with a Habitual Controller, a Goal-directed Controller, an Arbiter, and a 3D motion generation module. 
(iii) Extensive evaluation methods, human studies, and ablations demonstrate the effectiveness of GEMS and validate the benefits of introducing habitual behavior and multiple system coordination.

\section{Related work}

\paragraph{Human-like LLM agent.}
Human daily activity modeling supports applications in psychology, sociology, and virtual human simulation~\citep{garling1993psychological,adler1987everyday,schmidt2024frankenstein,fraser2024realistic}. Recent advances in LLMs have enabled increasingly realistic simulations of human behavior.
Existing works mainly follow two directions. One line models cognitive mechanisms behind behavior, including memory and planning~\citep{park2023generative}, desires~\citep{wang2023humanoid,wang2024d2a}, individual differences~\citep{he2024afspp,li2025x}, and social interactions~\citep{zhang2026context,lin2026asvo}. Another line grounds LLM-generated behaviors in 3D environments~\citep{leng2026agentsense,yu2024affordable,ye2026visually,zhang2025ella}. Most methods rely on predefined atomic actions, while ACTOR~\citep{liang2025actor} uses motion clips pre-generated by T2M models but may suffer from motion transition drift.
Despite these advances, existing frameworks are largely driven by explicit goals, plans or desires, following a goal-directed paradigm. In contrast, GEMS jointly models goal-directed and habitual behaviors and generates motions at runtime, enabling more diverse, natural, and believable behaviors in 3D environments.

\paragraph{Human behavior control mechanism study.}
An enduring and richly elaborated dichotomy in cognitive neuroscience is that of human behavior control mechanisms, divided into habitual and goal-directed~\citep{dolan2013goals,miller2019habits,wood2022habits}.
Existing computational frameworks of human behavioral control typically associate goal-directed control with model-based planning and habitual control with model-free learning~\citep{daw2005uncertainty,dolan2013goals,wood2016psychology}. Many works further introduce arbitration mechanisms that allocate control between these systems~\citep{daw2005uncertainty,lee2014neural}. Related cognitive architectures, including ACT-R, Soar, and the more recent CoALA, also employ procedural memory and model-free mechanisms related to habit learning for efficient action selection~\citep{taatgen2008acquisition,laird1987soar,sumers2024cognitive}.
However, most computational frameworks have focused on explaining multiple control systems from a classical reinforcement learning (RL) perspective, with limited exploration in realistic, open-ended 3D environments. On the other hand, such multiple systems accounts are well established in cognitive neuroscience and psychology, existing work on human-like agents has rarely explored such mechanisms.
To address this gap, we study multiple behavioral control system for human-like daily behavior simulation in realistic 3D environments.

\section{Methodology}

\subsection{Problem Setup}


\noindent\textbf{Human-like behavior simulation.}
Following prior works \citep{wang2024d2a,lin2026asvo}, we formulate human-like daily behavior simulation as a sequential decision-making problem \citep{littman1996algorithms}.
Formally, at each timestep $t$, a human-like agent observes a state $s_t$, which includes a psychological state component $s_t^{\mathrm{psy}}$. Multiple quantified factors are included in $s_t^{\mathrm{psy}}$, such as stress, depletion, and cognitive load.
We use $b_t$ to denote the agent's belief state, including its personality and memory.
The agent's goal $I$ can be specified by an external instruction or determined by the agent itself based on its current state $s_t$.
In this work, we primarily consider $I$ specified by external instructions.
The agent then generates an action $a_t^G$ based on its goal-directed system, i.e.,
\begin{equation}
a_t^G \sim \mathrm{Agent}(\cdot \mid s_t, b_t, I; p_{\theta}),
\end{equation}
where $p_\theta$ denotes the underlying LLM parameterized by $\theta$, and $a_t^G$ is generated to pursue the current behavioral goal $I$.
Following \citep{wang2024d2a}, each psychological factor of $s_t^{\mathrm{psy}}$ is represented on a $0$--$10$ Likert scale and updated by an LLM after each interaction based on the previous state, executed action, and environmental feedback.

\begin{figure}
    \centering
    \includegraphics[width=1\linewidth]{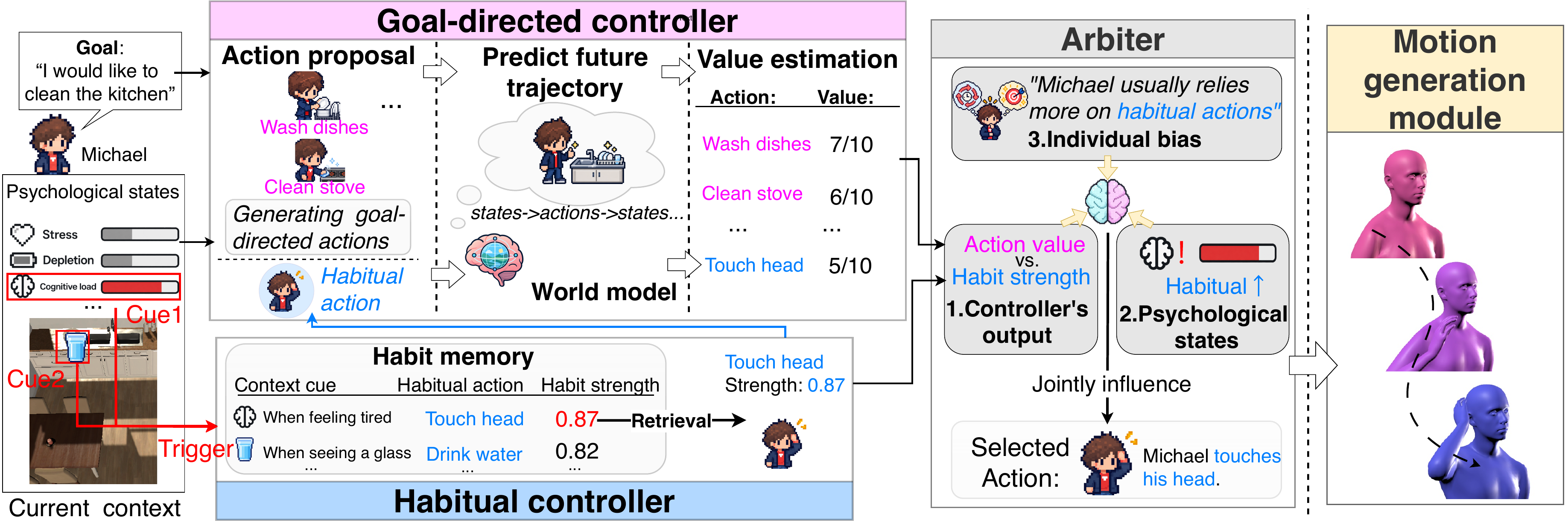}
    \caption{\textbf{Overall framework of the proposed GEMS.} The cognitive module takes the agent's goal and its current state as input: the Habitual Controller retrieves habitual actions based on cues triggered by the environment; the Goal-directed Controller proposes goal-directed actions and estimates the values of both habitual and goal-directed actions based on outcomes imagined by the world model; and the Arbiter selects an action based on both controller's output, individual bias, and psychological states. The selected action is then constructed by the motion generation module.}
    \label{fig:framework}
\end{figure}

\noindent\textbf{Human-like behavior simulation with multiple control systems.}
Unlike the conventional goal-directed setup, we extend behavior control to a multiple behavior control system formulation, where the agent can generate a goal-directed action $a_t^G$ and a habitual action $a_t^H$ under the goal-directed and habitual systems, respectively.
This enables the agent to make decisions under the joint influence of multiple control systems inspired by human behavioral control mechanisms.
The overall framework of the proposed GEMS is depicted in Fig. \ref{fig:framework}. The proposed GEMS consists of a \textbf{cognitive module} including Habitual Controller, Goal-directed Controller and Arbiter, as well as a \textbf{motion generation module} for human motion generation and object interaction in open-ended 3D environment.

\subsection{Cognitive module}

\subsubsection{Habitual controller}
Habitual behaviors can be triggered directly by a stimulus, i.e., \textit{a context cue} in the environment that activates the corresponding actions encoded in habit memory.
To simulate this process, we synthesize a prior habit dataset from empirical sources to support personalized habit memory creation for each agent, together with a cue retrieval algorithm. This easy-to-achieve design avoids the cost of modeling long-term habit formation.
At timestep $t$, the agent receives its current experienced context
$s_t$,
which may contain a context cue denoted as $q_i^{cue}$ that triggers a habitual action $a_i^H$.

\textbf{Habit memory.}
We construct the habit memory in three steps. 1) We model context cues. Potentially, any internal or external event may trigger habitual behavior~\citep{gardner2016habitual}. Since the full space of possible cues cannot be exhaustively represented, we represent the cue as a multidimensional vector whose dimensions are both empirically validated and available in the agent's current context $s_t$, including visual episodes, psychological states, events, and active goals. 
For simplicity, we assume that each habitual action is associated with a single cue dimension.
%
2) We synthesize a prior habit dataset from existing empirical sources, including habit studies and datasets \citep{ersche2017creature, moors2006automaticity}.
We extract habitual behaviors and their associated context cues from these sources and represent each instance in natural language as a cue--action pair $\langle q_i^{cue}, a_i^H \rangle$, with each cue mapped to a predefined cue dimension, yielding a habit dataset $\mathcal{D}^H$ containing 571 habitual behaviors observed in everyday human life. (see Appendix~\ref{app:habit_dataset} for details).
3) Finally, we initialize a personalized habit memory for each agent. Since individuals may exhibit different habits with varying strengths, we use LLMs to sample a subset of cue--action pairs from $\mathcal{D}^H$ based on the agent's persona and assign each sampled pair a habit strength $H(a_i^H)\in[0,1]$ (see Appendix~\ref{app:habit_initialization}). The resulting personalized habit memory can be written as $\widehat{\mathcal{D}}^H
=
\left\{
\left\langle
q_i^{\mathrm{cue}},
a_i^H, \right\rangle
H(a_i^H)
\right\}_{i=1}^{l}$ in which $l$ represents the number of cue-action pairs available for human-like agents.

\textbf{Cue retrieval algorithm.}
Habitual responses are triggered by recurring context cues, while stronger habits are more likely to guide behavior \citep{wood2016psychology}. Based on these observations, we propose a two-stage retrieval algorithm that first identifies relevant context cues from $\widehat{\mathcal{D}}^H$ given $s_t$, and then selects the most relevant habitual action in habit memory.
At the first stage, for each pair $\langle q_i^{cue}, a_i^H \rangle$ in $\widehat{\mathcal{D}}^H$, we compute the similarity $C_i$ between the corresponding cue dimension in $s_t$ and $q_i^{\mathrm{cue}}$.
We retain habits whose cue similarity exceeds a threshold $\kappa$, forming
$\mathcal{E}_t=\{i \mid C_i \geq \kappa\}$.
If $\mathcal{E}_t=\emptyset$, no habitual behavior is triggered and the agent relies solely on goal-directed control.
Otherwise, we proceed to the second stage and compute the response score for each habit $i\in\mathcal{E}_t$ as
$p_i=\frac{C_i-\kappa}{1-\kappa}H(a_i^H)$.
In step $t$, we denote the habitual action corresponding to the context cue $q_i^{cue}$ as $a_{i,t}^H$.
For $i\in\mathcal{E}_t$, we then select a habitual action according to the normalized distribution, i.e.,
$
P(a_{i,t}^H \mid s_t, \mathcal{E}_t)
=
\frac{p_i}
{\sum_{j\in\mathcal{E}_t} p_j}.
$
The two-stage design can reduce interference from strong but irrelevant habits while preserving the influence of habit strength on action selection, thereby supporting better habitual behavior generation.
We compare our method with alternative strategies on our designed evaluation set (see Appendix~\ref{app:habit_retrieval}) and it achieves the best performance (Tab.~\ref{tab:ablation_e}).

\subsubsection{Goal-Directed controller}
Goal-directed behaviors are driven by considering potential action outcomes, a process commonly modeled through model-based RL \citep{miller2019habits}.
To extend this principle to embodied environments with open-ended natural language action spaces, our Goal-directed Controller proposes candidate actions based on the agent's current state $s_t$ and goal $I$, predicts the consequences of these actions with an internal world model, and estimates their values for contributing to the goal. 

\textbf{Action proposal.} 
Denote $\mathcal{A}_t^G$ as the set of actions generated by the Goal-directed Controller at step $t$. 
The agent samples $N$ candidate actions that are likely to contribute positively to $I$, conditioned on $s_t$ and $b_t$, i.e., 
$\mathcal{A}_t^{G}=\{a_{i,t}^{G}\}_{i=1}^{N}\sim p_{\theta}(\cdot \mid s_t,b_t,I)$.
When a habitual action $a_{i,t}^H$ is triggered, it is considered alongside the goal-directed actions~\citep{morris2019generating}, yielding
$\mathcal{A}_t=\mathcal{A}_t^G\cup\{a_{i,t}^H\}$.

\textbf{Internal model.}
Humans rely on internal models to predict action consequences and adapt to dynamic environments~\citep{yu2026reinforcement}. Inspired by this mechanism, we introduce a frozen LLM-based world model~\citep{li2026bridging,nottingham2023embodied} $\mathcal{M}_{p_{\theta}}$ to approximate the environment transition function $T$. Given a candidate action $a_{i,t}\in\mathcal{A}_t$, the model predicts the next state as
\begin{equation}
\hat{s}_{t+1}
\sim
\mathcal{M}_{p_{\theta}}(\cdot\mid s_t,a_{i,t})
\approx
T(\cdot\mid s_t,a_{i,t}).
\end{equation}
By recursively applying the world model, the agent can prospectively predict its future states following each candidate action. Further design details of the world model are provided in Appendix~\ref{world_model}.

\textbf{Value estimation.}
The internal model enables the agent to estimate the goal-directed value of each candidate action through prospective planning. Starting from $a_{i,t}\in\mathcal{A}_t$, a planning strategy $\mathcal{P}$ simulates possible future developments and evaluates the imagined outcomes with respect to $I$:
\begin{equation}
Q(s_t,a_{i,t}\mid I)
=
\mathbb{E}_{\hat{\tau}\sim
\mathcal{P}(\mathcal{M}_{p_{\theta}};s_t,a_{i,t},I)}
\left[
\mathcal{V}_{p_{\theta}}(\hat{\tau},I)
\right],
\end{equation}
where $\hat{\tau}$ denotes an imagined future trajectory and $\mathcal{V}_{p_{\theta}}$ prompts the agent to assign a subjective score from 0 to 10 according to how well the predicted outcome satisfies $I$.
For simplicity, we write $Q(a_{i,t})$ for $Q(s_t,a_{i,t}\mid I)$ hereafter.
%

\subsubsection{Arbiter}
\label{arbiter}
The Goal-directed and Habitual Controllers may favor different actions under the same experienced context $s_t$. Additionally, considering individual differences and their state (i.e., environment and psychological state) variations of real human, 
our Arbiter jointly integrates three sources of influence: the output of each controller~\citep{miller2019habits}, the agent's individual control bias~\citep{piray2016human}, and its current psychological state~\citep{wood2022habits}, for selecting the final action. This mechanism is motivated by findings in psychology and neuroscience and shows how multiple behavior control systems jointly shape the final behavior. We represent each influence as:

\textit{1) Controllers' output}: For an action $a_{i,t}$, the Goal-directed Controller provides action value $Q(a_{i,t})$, while the Habitual Controller provides habit strength $H(a_{i,t})$. For actions proposed only by the Goal-directed Controller, we set $H(a_{i,t})=0$. A higher output value indicates stronger influence from the corresponding controller.
\textit{2) Individual control bias}: We further denote $\rho\in(0,1)$ as the agent's baseline preference for goal-directed control, where larger values indicate that the agent relies more on goal-directed control.
%
\textit{3) Current psychological state}: 
We design a psychological state interference function
$F(s_t^{\mathrm{psy}})\in[0,1]$ to characterize how adverse psychological states increase reliance on habitual control.
For each psychological factor in $s_t^{\mathrm{psy}}$, quantified on a 0--10 scale, A higher score indicates a more negative psychological state, making $F(s_t^{\mathrm{psy}})$ more likely to take a larger value and thereby increasing the weight of Habitual Controller. Here, we consider three psychological factors: stress (\textit{str.}), depletion (\textit{dep.}) and cognitive load (\textit{cog.}) \citep{wood2016psychology}.
$F(s_t^{\mathrm{psy}})$ introduces transient dynamics induced by the agent's current psychological state into the balance with the two controllers. It can be instantiated in various ways, such as using an LLM. In this work, we calibrate $F(s_t^{\mathrm{psy}})$ using estimation values from existing human behavioral studies.
More details are provided in Appendix~\ref{gate}.

Inspired by the arbitration formulation in \citet{miller2019habits}, we further define the overall drive \(U(a_{i,t})\) of each candidate action by considering the three sources of influence together:
\begin{equation}
U(a_{i,t})
=
w_t^G \bar{Q}(a_{i,t})
+
w_t^H H(a_{i,t}),
\end{equation}
where \(w_t^G = \rho\left(1-F(s_t^{\mathrm{psy}})\right)\) denotes the joint weight of goal-directed control at time \(t\), and the corresponding habitual control weight is given by \(w_t^H = 1-w_t^G\), and \(\bar{Q}(a_{i,t}) \in [0,1]\) denotes the rescaled goal-directed value of \(Q(a_{i,t})\).
Finally, the Arbiter selects the action with the highest overall drive as the final action of the human-like agent, i.e., 
$
a_t
=
\arg\max_{a_{i,t} \in \mathcal{A}_t} U(a_{i,t}).
$
This design accounts for the intrinsic action preferences of each controller while incorporating individual differences and dynamic psychological states, thereby supporting heterogeneous decision-making and promoting coherent and natural behavior sequences.

\begin{figure}
    \centering
    \includegraphics[width=1\linewidth]{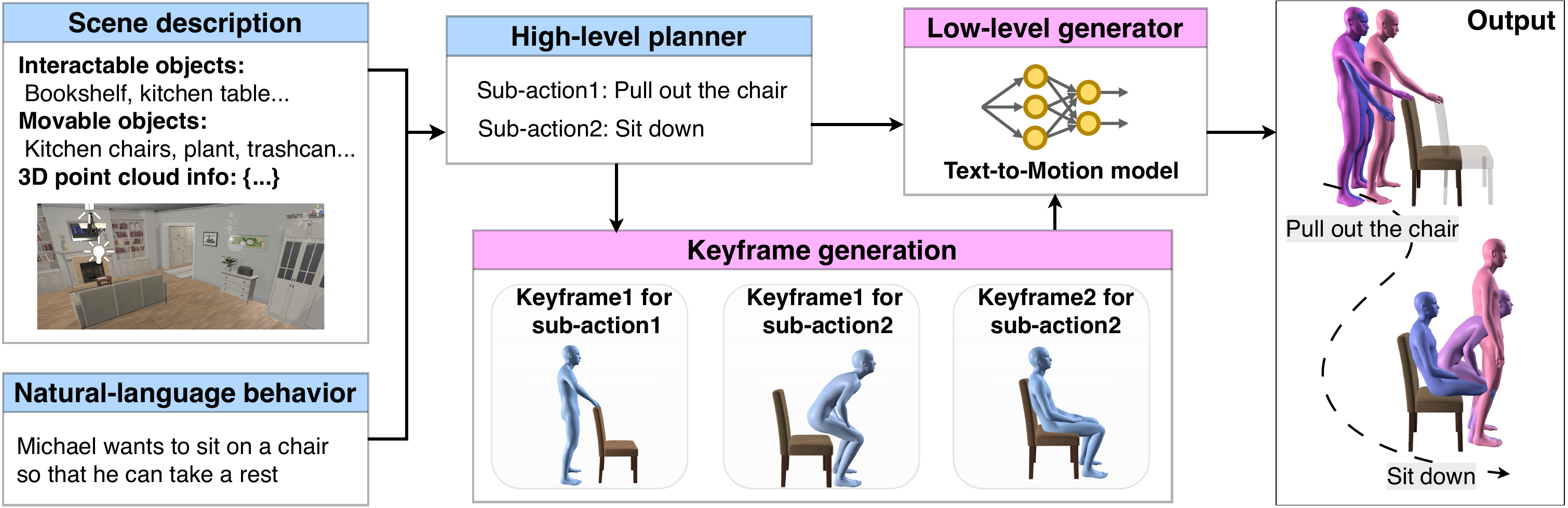}
    \caption{The motion generation module consists of two components. Given the action description in natural language from the cognitive module and the current scene information, the high-level planner produces an executable sub-action plan; sub-actions are aligned with keyframes where needed. Using this as a condition, the low-level generator uses T2M models to generate motion sequences.}
    \label{fig: motion_generation_module}
    \vspace{-0.45cm}
\end{figure}

\subsection{Motion generation module}
Unlike existing approaches~\citep{liang2025actor,leng2026agentsense} constrained by a predefined action space, our motion generation module constructs human–scene interactions in 3D environments from the natural language action $a_t$. It employs a High-Level Planner and a Low-Level Motion Generator.

\textbf{High-Level Planner.}
As shown in Fig.~\ref{fig: motion_generation_module}, the planner augments the action description $a_t$ with necessary execution steps required to achieve the action in the current scene. Given $a_t$ and the current scene state, including the scene layout, objects, and the agent's states, the planner determines the desired scene state after the action is completed and, when necessary, decomposes the action into an ordered sequence of executable sub-actions and object interactions. For instance, if the agent wants to sit on a chair, the chair should be pulled out before sitting down. An LLM is used to produce this ordered decomposition, yielding a sequence of textual sub-action descriptions and matched keyframe constraints used to guide the Low-Level Motion Generator.

\textbf{Low-Level Motion Generator.}
Given the decomposed sub-action descriptions, the generator executes them sequentially by synthesizing human motion segments and coordinating object interactions, yielding diverse human-object interaction sequences.
In particular, the scene-agnostic T2M model~\citep{rempe2026kimodo} we adopt offers expressive motion synthesis but does not directly perceive scene geometry. To address this limitation, we introduce a keyframe mechanism that bridges high-level planning and low-level motion generation by encoding planned sub-actions and scene information as 3D constraints for motion synthesis~\citep{xie2024omnicontrol}. Given a sub-action instruction and the perceived scene point cloud, a VLM predicts target human skeletal poses and corresponding object poses for a sparse set of keyframes. This mechanism supports flexible sub-action composition and smooth motion transitions.
The T2M model then synthesizes each motion segment from the sub-action description, incorporating keyframe constraints and a planned navigation path when applicable. When the generated motion reaches a designated keyframe, the system triggers the corresponding interaction event, such as establishing a hand-object attachment when grasping an apple.
Further implementation details are provided in Appendix~\ref{3Dmodule}.

\begin{table*}[t]
\centering
\caption{Quantitative comparison with previous human-like agent frameworks on controlled daily-life goals with the same initialization. Mean denotes the average of Nat., Coh., and PA. Besides, Qual. and Act. have a maximum score of 5. The best results are highlighted in \textbf{bold}.}
\label{tab:major_performance}
\vspace{1mm}
\scriptsize
\renewcommand{\arraystretch}{1.08}
\setlength{\tabcolsep}{4pt}
\begin{tabular}{@{}lcccccccccc@{}}
\toprule
\multirow{2}{*}{Method}
& \multicolumn{4}{c}{Cognitive module}
& \multicolumn{5}{c}{Motion generation module}
& \multirow{2}{*}{Action type} \\
\cmidrule(lr){2-5}\cmidrule(lr){6-10}
& Nat. $\uparrow$ & Coh. $\uparrow$ & PA $\uparrow$ & Mean. $\uparrow$
& SSR $\uparrow$ & GSR $\uparrow$ & DTG $\downarrow$ & Qual. $\uparrow$ & Act. $\uparrow$ & \\
\midrule

X-VirtualHome~{\scriptsize\citep{leng2026agentsense}}
& 42.26 & 83.81 & 77.52 & 67.86
& 56.67 & 53.33 & \textbf{19.95} & 1.27 & 2.86 & Atomic \\

AGA~{\scriptsize\citep{yu2024affordable}}
& 56.98 & 53.74 & 69.17 & 59.96
& 51.33 & 26.67 & 19.96 & 1.21 & 1.68 & Atomic \\

D2A~{\scriptsize\citep{wang2024d2a}}
& 63.74 & 76.90 & 75.43 & 72.02
& -- & -- & -- & -- & -- & -- \\

ASVO~{\scriptsize\citep{lin2026asvo}}
& 42.01 & 46.83 & 27.11 & 38.65
& -- & -- & -- & -- & -- & -- \\

ACTOR~{\scriptsize\citep{liang2025actor}}
& 47.77 & 86.18 & 77.62 & 70.52
& 59.79 & 46.67 & 98.42 & 3.75 & 3.59 & Motion clips \\

ACTOR$^\dagger$~{\scriptsize\citep{liang2025actor}}
& 52.79 & 85.85 & 78.27 & 72.30
& 62.67 & 48.67 & 94.74 & 3.66 & 3.69 & Motion clips \\

\midrule
GEMS (ours)
& \textbf{90.31} & \textbf{89.96} & \textbf{84.35} & \textbf{88.21}
& \textbf{86.24} & \textbf{79.33} & 22.48 & \textbf{4.12} & \textbf{4.23} & \textbf{Generative} \\
\bottomrule
\end{tabular}
\end{table*}

\section{Experiment}

\subsection{Experimental setup and evaluation metrics}

To evaluate the human-like performance of GEMS, we separately evaluate its two main modules:
(i) the cognitive module, by assessing the human-likeness of the behavior sequences it generates; and
(ii) the motion generation module, by assessing whether the selected behaviors can be faithfully and naturally constructed in 3D environments.
For the cognitive module, we follow prior LLM-based evaluation protocols~\citep{wang2024d2a} and use three metrics, each scored on a 0--100 scale:
\textbf{Naturalness (Nat.)} for human-likeness~\citep{wang2024d2a},
\textbf{Coherence (Coh.)} for logical consistency~\citep{wang2024d2a}, and
\textbf{Personality Alignment (PA)} for consistency with the given persona~\citep{li2025x}.
\textbf{Mean} denotes their average value. This evaluation protocol emulates
a human user study \citep{cai2025diffusion}.
For the motion generation module, \textbf{Step Success Rate (SSR)} measures individual-step success, and \textbf{Goal Success Rate (GSR)} measures whole-plan success \citep{liang2025actor}. 
\textbf{Distance to Goal (DTG)} measures the minimum wrist-to-goal-object distance~\citep{diomataris2024wandr}. 
\textbf{Quality Score (Qual.)} measures motion naturalness, while 
\textbf{Action Score (Act.)} measures consistency with the target behavior description~\citep{wang2022humanise}.
The evaluation criteria and the definitions of successful task execution follow the same settings in the corresponding prior works. Details are provided in Appendix \ref{metric_details}.

We compare GEMS with five previous human-like agent frameworks.
X-VirtualHome, AGA, and ACTOR, which support embodied behavior execution and can therefore be evaluated on both the cognitive and motion generation module metrics, whereas D2A and ASVO only focus on language behavior generation and are therefore evaluated only on the cognitive module metrics.
We further evaluate a variant of ACTOR, denoted as ACTOR$^\dagger$, which retains the original ACTOR pipeline while using the same Habit Memory (with habitual behaviors sampled using an LLM), T2M model, and simulation platform as GEMS.
For a fair comparison, we evaluate all methods on 30 shared daily life goals (such as preparing dinner and cleaning the kitchen) with identical initialization settings.
Each goal runs five times, yielding 150 runs per method, and all scores are averaged over runs. 
Full implementation details are provided in Appendix~\ref{implementation_details}.

\begin{figure}
    \centering
    \includegraphics[width=1\linewidth]{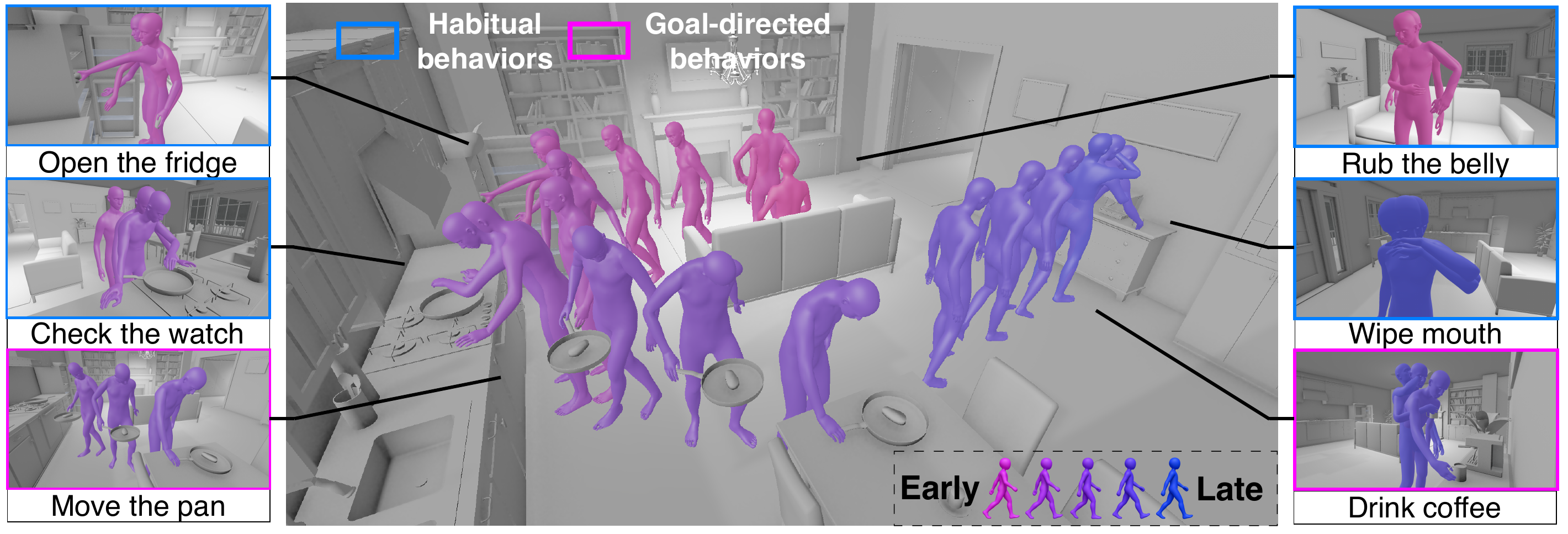}
    \caption{Qualitative example generated by GEMS for the goal ``prepare dinner''.} 
    \label{fig:qualitative_showcase}
\end{figure}

\subsection{Comparison with state-of-the-art methods}
\label{quantitativeevaluation}

As shown in Tab.~\ref{tab:major_performance}, ACTOR$^\dagger$ yields only limited gains over ACTOR, suggesting that the effective triggering and arbitration of habitual behaviors are critical, thereby highlighting the importance of the Habitual Controller and Arbiter design. The limited improvement in motion metrics further emphasizes the importance of keyframe-guided motion generation.
In contrast, GEMS outperforms all the baselines on the cognitive module metrics, and improves the Mean score by about 16 points over the previous best method D2A.
These results demonstrate the benefit of modeling habitual behaviors and coordinating habitual and goal-directed control for believable human-like behavior.
For the motion generation module, GEMS also demonstrates the best overall performance.
Compared with predefined atomic actions or independent motion clips, our keyframe-guided approach generates continuous, scene-aware motions at runtime. This improves the overall quality of the generated motions.
A qualitative visualization of GEMS are shown in Fig.~\ref{fig:qualitative_showcase}, with more results in Appendix~\ref{case_study2}.

\begin{wrapfigure}{r}{0.5\linewidth}
    \centering
    \vspace{-8pt}
    \includegraphics[width=\linewidth]{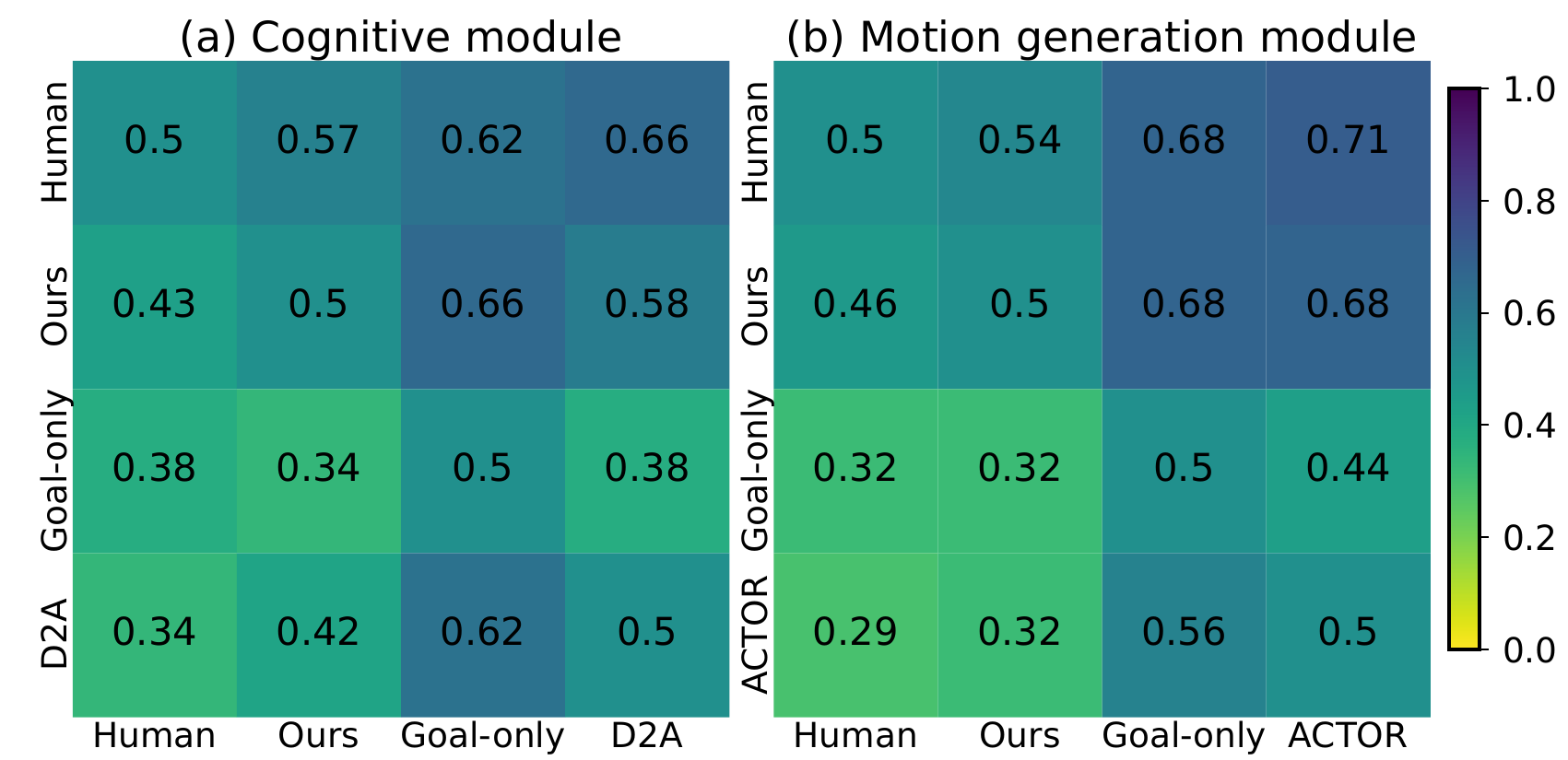}
    \caption{
        Key human preference results. Each entry compares the win rate of the method on the vertical axis against that on the horizontal axis, with diagonal values fixed at 0.5. Full results are provided in Appendix~\ref{humanstudy}.
    }
    \label{fig:human_man}
    \vspace{-8pt}
\end{wrapfigure}

\subsection{Human study}

To evaluate whether incorporating habitual behavior improves perceived human-likeness and to assess how closely GEMS approximates real human behavior, we conduct blind pairwise human evaluations through questionnaires. We evaluate the language behavioral sequences generated by the cognitive module and the motion videos generated by the motion generation module, with a total of \textbf{94} crowd-sourced participants involved in the study.

We further compare GEMS with a variant that uses only the Goal-directed Controller (denoted as Goal-only), other baseline methods, and real human behavior references~\citep{yang2025egolife,jiang2024scaling}. In each pairwise comparison, participants are asked to select the behavior that appears more human-like.
As shown in Fig.~\ref{fig:human_man}, GEMS receives higher preference rates than Goal-only, demonstrating that incorporating habitual behavior improves perceived human-likeness. When directly compared with real human behavior, GEMS achieves preference rates of 43.4\% and 46.3\% in the two evaluations, respectively, indicating that GEMS is capable of generating believable and human-like behaviors.
Among all agent frameworks, GEMS achieves the highest overall preference rate. The complete study design and detailed results are provided in Appendix~\ref{humanstudy}.

\begin{table*}[t]
\centering
\caption{\textbf{Ablation study.}
(a) Ablation of the key components in the cognitive module.
(b) Ablation of $F(s_t^{\mathrm{psy}})$ in the Arbiter under different initializations of $s_t^{\mathrm{psy}}$.
Initial states $s_0^{\mathrm{psy}}$ include (stress, depletion, cognitive load): \textit{Normal} uses \((3,3,3)\); \textit{Adverse} denotes negative psychological states and it averages results across three settings: \((7,3,3)\), \((3,7,3)\), and \((3,3,7)\).
(c) Comparison of alternative motion-generation methods.
(d) Ablation of keyframe guidance in the motion generation module.
(e) Comparison of different habit retrieval methods.
(f) Effect of the coefficient $\rho$.
Habit Ratio measures the proportion of habitual actions, while TE measures task execution efficiency.}
\label{tab:ablation_all}
\vspace{-1mm}
\scriptsize
\renewcommand{\arraystretch}{0.95}
\captionsetup[subtable]{skip=0pt}
\setlength{\abovecaptionskip}{1pt}
\setlength{\belowcaptionskip}{0pt}

\noindent
\begin{subtable}[t]{0.30\linewidth}
\centering
\setlength{\tabcolsep}{1.8pt}
\begin{tabular}{@{}ccc|ccc@{}}
\toprule
Habit & WM & Arb. & Nat. $\uparrow$ & Coh. $\uparrow$ & PA $\uparrow$ \\
\midrule
& & & 39.54 & 55.39 & 71.48 \\
\checkmark & & & 61.05 & 74.72 & 81.80 \\
& \checkmark & & 52.25 & 84.66 & 73.92 \\
\checkmark & \checkmark & & 76.17 & 72.67 & 82.83 \\
\checkmark & \checkmark & \checkmark & \textbf{88.48} & \textbf{87.38} & \textbf{85.22} \\
\bottomrule
\end{tabular}
\caption{}\label{tab:ablation_a}
\end{subtable}%
\hfill
\begin{subtable}[t]{0.25\linewidth}
\centering
\setlength{\tabcolsep}{1.2pt}
\begin{tabular}{@{}c|ccc@{}}
\toprule
$F(s_t^{\mathrm{psy}})$ & \multicolumn{2}{c}{Habit Ratio (\%)} & Mean. $\uparrow$ \\
\cmidrule(lr){2-3}
& Normal & Adverse  & \\
\midrule
& 35.81 & 36.57 & 79.54 \\
\checkmark & 40.92 & \textbf{53.06} & \textbf{85.71} \\
\bottomrule
\end{tabular}
\caption{}\label{tab:ablation_b}
\end{subtable}%
\hfill
\begin{subtable}[t]{0.41\linewidth}
\centering
\setlength{\tabcolsep}{1.6pt}
\begin{tabular}{@{}lccc@{}}
\toprule
Motion generation method & SSR $\uparrow$ & Qual. $\uparrow$ & Act. $\uparrow$ \\
\midrule
HUMANISE~{\scriptsize\citep{wang2022humanise}} & 33.33 & 1.38 & 1.31 \\
HOI-from-HLI~{\scriptsize\citep{wu2025human}} & 37.78 & 2.13 & 1.72 \\
InfBaGel~{\scriptsize\citep{zou2026infbagel}} & 40.00 & 1.76 & 1.69 \\
GEMS (ours) & \textbf{95.56} & \textbf{4.23} & \textbf{4.52} \\
\bottomrule
\end{tabular}
\caption{}\label{tab:ablation_c}
\end{subtable}

\par\vspace{-0.5mm}\noindent
\begin{subtable}[t]{0.26\linewidth}
\centering
\setlength{\tabcolsep}{1.5pt}
\begin{tabular}{@{}c|ccc@{}}
\toprule
Keyframe & SSR $\uparrow$ & Qual. $\uparrow$ & DTG $\downarrow$ \\
\midrule
& 52.22 & 1.75 & 76.80 \\
\checkmark & \textbf{86.24} & \textbf{4.12} & \textbf{22.48} \\
\bottomrule
\end{tabular}
\caption{}\label{tab:ablation_d}
\end{subtable}%
\hfill
\begin{subtable}[t]{0.25\linewidth}
\centering
\setlength{\tabcolsep}{1.2pt}
\begin{tabular}{@{}l|cc@{}}
\toprule
Retrieval method &  Acc. (\%) $\uparrow$ \\
\midrule
Sim.$\times$HS & 71.4 \\
Sim.-only & 73.3 \\
GEMS (ours) & \textbf{84.7} \\
\bottomrule
\end{tabular}
\caption{}\label{tab:ablation_e}
\end{subtable}%
\hfill
\begin{subtable}[t]{0.45\linewidth}
\centering
\setlength{\tabcolsep}{1.4pt}
\begin{tabular}{@{}l|ccccccc@{}}
\toprule
$\rho$ & 0.05 & 0.1 & 0.3 & 0.5 & 0.7 & 0.9 & 0.95 \\
\midrule
Habit Ratio (\%) & 55.62 & 52.27 & 52.94 & 41.83 & 41.65 & 9.46 & 9.92 \\
TE $\uparrow$ & 0.51 & 0.59 & 0.57 & 0.78 & 0.66 & \textbf{0.99} & 0.81 \\
Mean. $\uparrow$ & 77.37 & 80.53 & 85.87 & \textbf{87.31} & 84.74 & 78.00 & 75.53 \\
\bottomrule
\end{tabular}
\caption{}\label{tab:ablation_f}
\end{subtable}

\vspace{-1mm}
\end{table*}


\subsection{Ablation study}
\label{ablation}

We define several additional metrics for the ablation studies. For a completed trajectory $\tau$ consisting of $m$ execution steps,
\textbf{Habit Ratio} is the fraction of habitual actions, and 
\textbf{Task Execution Efficiency (TE)} is 
$\mathrm{TE}=\frac{\mathcal{V}_{p_\theta}(\tau,I)}{m}$.
\textbf{Retrieval Accuracy (Acc.)} measures correctness on the retrieval evaluation set (see Appendix~\ref{app:retrieval_algorithm}).
Each setting in Tab.~\ref{tab:ablation_a}, \ref{tab:ablation_b}, and \ref{tab:ablation_f} is \textit{averaged} over 15 independent runs.

\noindent\textbf{Impact of the Key Components in the Cognitive Module.}
Tab.~\ref{tab:ablation_a} evaluates the effectiveness of the Habitual Controller, World Model, and Arbiter.
Without the Arbiter, an LLM selects among the actions proposed by two controllers; without the World Model, an LLM directly estimates action values.
Adding the Habitual Controller improves Nat., supporting our motivation to model habitual behaviors for believable human-like agents simulation.
Adding the Arbiter further improves all three metrics, suggesting that generating natural and coherent human-like behavior benefits not only from incorporating habitual behaviors but also from coordinating habitual and goal-directed control.

\noindent\textbf{Ablation on $F(s_t^{\mathrm{psy}})$ in Arbiter.}
Tab.~\ref{tab:ablation_b} evaluates the effect of $F(s_t^{\mathrm{psy}})$ under different psychological state initializations.
\textit{Normal} sets the initial psychological state (stress, depletion, cognitive load) to $s_0^{\mathrm{psy}}=(3,3,3)$.
\textit{Adverse} averages the results across three settings: $(7,3,3)$, $(3,7,3)$, and $(3,3,7)$.
Adding $F(s_t^{\mathrm{psy}})$, the \textit{Adverse} setting increases the habit ratio, while this shift is attenuated without it. This shows that $F(s_t^{\mathrm{psy}})$ enables the Arbiter to dynamically balance two controllers according to psychological states (see Tab.~\ref{tab:ablation_gate} for detailed results).

\noindent\textbf{Comparison of the Motion Generation Module.}
Tab.~\ref{tab:ablation_c} compares alternative human–scene interaction generation methods for our motion generation module on 45 randomly sampled actions generated by the cognitive module.
Existing methods achieve low SSR, highlighting their limitation in constructing the diverse behaviors in our setting and motivating the design of our module.

\noindent\textbf{Impact of Keyframe Guidance.}
Tab.~\ref{tab:ablation_d} evaluates keyframe guidance in the motion generation module.
Introducing keyframes improves SSR and motion quality while reducing DTG, showing that keyframe constraints effectively ground the scene-agnostic T2M model in 3D interactions while improving overall motion naturalness.

\noindent\textbf{Comparison of Retrieval Methods.}
Tab.~\ref{tab:ablation_e} compares our two-stage retrieval with cue similarity alone (Sim.-only) and cue similarity multiplied by habit strength (Sim.$\times$HS).
Our two-stage method achieves the best performance, supporting the separation of cue matching from habit-strength weighting (see Tab.~\ref{tab:retrieval_threshold_test} for detailed results).
The results in Tab.~\ref{tab:ablation_a} further suggest that effective habit retrieval schema in the Habit Controller contributes to the human-like performance.

\noindent\textbf{Influence of the Individual Bias Coefficient $\rho$.}
Tab.~\ref{tab:ablation_f} studies the influence of $\rho$.
Different values of $\rho$ induce distinct behavior patterns, showing that the Arbiter provides a controllable mechanism for individual variation.
Increasing $\rho$ shifts behavior control toward goal-directed behavior, reducing the habit ratio while generally improving TE.
An intermediate value of $\rho$ achieves the best Mean score, indicating a trade-off between habit ratio and TE in achieving human-like performance. Based on these results, we recommend \(\rho \in [0.3, 0.7]\) as the preferred range for persona configuration. These findings suggest that simply increasing the habit ratio does not necessarily yield more human-like performance; rather, effective coordination between the two controllers is the key.

\section{Conclusion}
In this work, we propose GEMS, a Generative Embodied Multiple behavioral control Systems for building human-like agent by jointly modelling goal-directed and habitual behaviors through a Habitual Controller, a Goal-directed Controller, and an Arbiter, together with a keyframe-guided motion generation module for converting natural language behavior descriptions into 3D human-object interactions. Extensive quantitative evaluations and human studies demonstrate that GEMS generates more believable human-like behaviors. Further ablation studies validate the benefits of multiple behavioral control system coordination, and keyframe-guided motion generation. We hope this work represents a step toward more natural and diverse human-like agents in interactive 3D environments and offers valuable insights for future research.

\section*{AI Use Disclosure}
In this work, we used generative AI for data cleaning and reformatting, as well as for synthetic data generation. In addition, we used ChatGPT for automated quantitative evaluation. We did not use generative AI tools for research idea development, experimental design, or the analysis of experimental results. All AI-assisted work was reviewed or validated by the authors. The authors take full responsibility for the final content of this work.

\section*{Ethics Statement}

This work includes human-subject evaluations conducted to assess the perceived
human-likeness of generated behavioral sequences and 3D human motions.
The study protocol was reviewed and approved by the ethics committee of our
institution prior to participant recruitment.
Participants were recruited through Amazon Mechanical Turk (MTurk), and all
participants provided informed consent before taking part in the study.
Participation was limited to individuals aged 18 years or older.
We collected demographic information, including age, gender, ethnicity, and
education level, solely for descriptive characterization of the participant
population.
All responses were analyzed in de-identified form, and the collected data were
used solely for the human evaluations reported in this work.
Further details of the study protocol, participant population, and evaluation
procedure are provided in Appendix~\ref{humanstudy}.


\bibliography{iclr2027_conference}
\bibliographystyle{iclr2027_conference}

\clearpage

\appendix

\section{Experiment details}

\begin{table*}[t]
\centering
\caption{Quantitative comparison with previous human-like agent frameworks under long-horizon daily simulation.}
\label{tab:long_horizom__performance}
\vspace{1mm}
\scriptsize
\renewcommand{\arraystretch}{1.08}
\setlength{\tabcolsep}{4pt}
\begin{tabular}{@{}lcccccccccc@{}}
\toprule
\multirow{2}{*}{Method}
& \multicolumn{4}{c}{Cognitive module}
& \multicolumn{5}{c}{Motion generation module}
& \multirow{2}{*}{Action type} \\
\cmidrule(lr){2-5}\cmidrule(lr){6-10}
& Nat. $\uparrow$
& Coh. $\uparrow$
& PA $\uparrow$
& Mean. $\uparrow$
& SSR $\uparrow$
& GSR $\uparrow$
& DTG $\downarrow$
& Qual. $\uparrow$
& Act. $\uparrow$
& \\
\midrule

X-VirtualHome~{\scriptsize\citep{leng2026agentsense}}
& 45.18 & 84.62 & 78.10 & 69.30
& 58.11 & 54.22 & \textbf{19.73} & 1.31 & 2.91
& Atomic \\

AGA~{\scriptsize\citep{yu2024affordable}}
& 59.42 & 55.87 & 70.31 & 61.87
& 52.40 & 28.11 & 19.81 & 1.24 & 1.71
& Atomic \\

ACTOR~{\scriptsize\citep{liang2025actor}}
& 53.84 & 87.21 & 79.05 & 73.37
& 61.08 & 47.15 & 96.31 & 3.78 & 3.62
& Motion clips \\

ACTOR$^\dagger$~{\scriptsize\citep{liang2025actor}}
& 57.12 & 87.66 & 80.13 & 74.97
& 64.02 & 50.21 & 91.76 & 3.70 & 3.73
& Motion clips \\

\midrule
\textbf{Ours}
& \textbf{87.19}
& \textbf{88.34}
& \textbf{85.01}
& \textbf{86.85}
& \textbf{79.11}
& \textbf{70.24}
& 21.87
& \textbf{4.15}
& \textbf{4.25}
& \textbf{Generative} \\

\bottomrule
\end{tabular}
\end{table*}

\subsection{Additional ablation studies}

\begin{table*}[t]
\centering
\caption{\textbf{Additional ablation studies.}
(a) Comparison of different planning algorithms.
(b) Comparison of different LLMs, with GPT-5.5 used by default.
(c) Effect of $F(s_t^{\mathrm{psy}})$ under different psychological states with $\rho=0.5$.}
\label{tab:additional_ablation}
\vspace{-1mm}

\scriptsize
\renewcommand{\arraystretch}{0.95}
\captionsetup[subtable]{skip=1pt}
\setlength{\abovecaptionskip}{1pt}
\setlength{\belowcaptionskip}{0pt}

\begin{subtable}[t]{0.19\linewidth}
\centering
\resizebox{\linewidth}{!}{%
\begin{tabular}{@{}lcc@{}}
\toprule
Algorithm & TE $\uparrow$ & Mean $\uparrow$ \\
\midrule
Greedy & 0.69 & \textbf{89.97} \\
Beam   & 0.71 & 88.23 \\
MCTS   & \textbf{0.79} & 88.19 \\
\bottomrule
\end{tabular}}
\caption{}
\label{tab:ablation_planning}
\end{subtable}
\hfill
\begin{subtable}[t]{0.30\linewidth}
\centering
\resizebox{\linewidth}{!}{%
\begin{tabular}{@{}lccc@{}}
\toprule
LLM & Mean $\uparrow$ & SSR $\uparrow$ & GSR $\uparrow$ \\
\midrule
DeepSeek v4.1 Flash & \textbf{89.33} & 86.47 & 80.00 \\
GPT-5.5             & 88.37 & 88.12 & \textbf{86.67} \\
GPT-5.6             & 87.09 & \textbf{89.95} & 73.33 \\
\bottomrule
\end{tabular}}
\caption{}
\label{tab:ablation_llm}
\end{subtable}
\hfill
\begin{subtable}[t]{0.49\linewidth}
\centering
\resizebox{\linewidth}{!}{%
\begin{tabular}{@{}llccc@{}}
\toprule
Condition
& $(\textit{str.},\textit{dep.},\textit{cog.})$
& $F(s_t^{\mathrm{psy}})$
& Habit Ratio (\%)
& Mean $\uparrow$ \\
\midrule
\multirow{2}{*}{Stress}
& \multirow{2}{*}{$(7,3,3)$}
& & 38.76 & 76.48 \\
& & $\checkmark$ & 50.84 & 87.56 \\
\addlinespace

\multirow{2}{*}{Depletion}
& \multirow{2}{*}{$(3,7,3)$}
& & 31.80 & 80.37 \\
& & $\checkmark$ & 52.31 & 82.90 \\
\addlinespace

\multirow{2}{*}{Cog. Load}
& \multirow{2}{*}{$(3,3,7)$}
& & 39.15 & 79.39 \\
& & $\checkmark$ & 56.03 & 88.04 \\
\addlinespace

\multirow{2}{*}{Normal}
& \multirow{2}{*}{$(3,3,3)$}
& & 35.81 & 81.92 \\
& & $\checkmark$ & 40.92 & 84.34 \\
\midrule

Ref. from Tab.~\ref{tab:ablation_f}
& $(3,3,3)$
& $\rho=0.05$
& 55.62 & 77.37 \\
\bottomrule
\end{tabular}}
\caption{}
\label{tab:ablation_gate}
\end{subtable}

\vspace{-1mm}
\end{table*}

\noindent\textbf{Comparison of Planning Algorithms.}
Tab.~\ref{tab:ablation_planning} compares different planning algorithms for the planning procedure $\mathcal{P}$. Different planning strategies can instantiate $\mathcal{P}$, including greedy search~\citep{resende2013grasp} and Monte Carlo tree search (MCTS)~\citep{swiechowski2023monte}.
MCTS achieves the highest TE, while Greedy obtains the highest Mean score. The three planning algorithms achieve comparable Mean scores, indicating that GEMS is relatively robust to the choice of planning strategy. Meanwhile, the higher TE achieved by MCTS suggests that stronger prospective search can improve goal-directed task execution efficiency. 

\noindent\textbf{Comparison of LLMs.}
Tab.~\ref{tab:ablation_llm} evaluates GEMS with different LLM backbones. DeepSeek achieves the highest Mean score, GPT-5.6 achieves the highest SSR, and GPT-5.5 achieves the highest GSR. The three LLM backbones yield comparable overall performance across the cognitive and motion-generation metrics, suggesting that the effectiveness of GEMS is not specific to a particular LLM backbone. We use GPT-5.5 as the default LLM in our experiments.

\noindent\textbf{Effect of $F(s_t^{\mathrm{psy}})$.}
Tab.~\ref{tab:ablation_gate} further examines the effect of the psychological state interference function $F(s_t^{\mathrm{psy}})$ under different psychological states. Under each psychological-state initialization, enabling
$F(s_t^{\mathrm{psy}})$ yields both a higher habit ratio and
a higher Mean score than its ablated counterpart. These paired
ablations support the effectiveness of $F(s_t^{\mathrm{psy}})$
in strengthening state-dependent habitual responding and
improving evaluated behavioral quality. As a complementary
reference, the adverse cognitive-load setting achieves a higher
Mean score than the low-$\rho$ setting in
Tab.~\ref{tab:ablation_f}, despite their similar habit ratios
of around 55\%. This comparison further suggests that overall
habit frequency alone does not adequately characterize the
quality of generated behavior.

\begin{figure}
    \centering
    \includegraphics[width=1\linewidth]{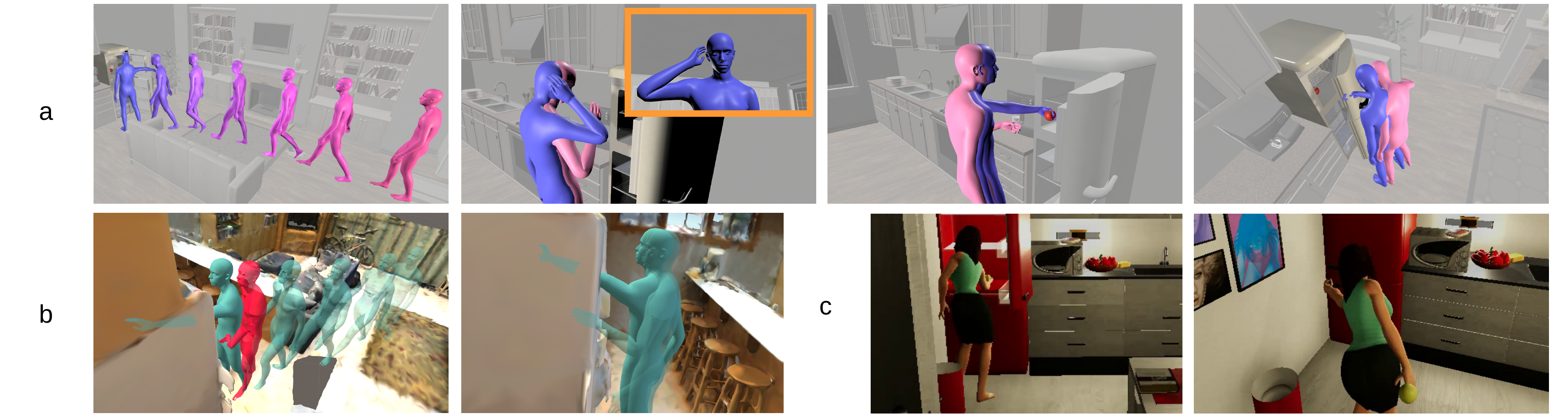}
    \caption{Qualitative comparison under the goal ``find something to eat in the fridge.'' We compare (a) Ours, (b) ACTOR, and (c) X-VirtualHome. While (b)--(c) are largely task-oriented, ours exhibits richer habitual behaviors and appears more human-like. \textcolor{red}{Red} frames indicate skeletal drift.}
    \label{fig:qualitative_comparison}
\end{figure}

\subsection{Implementation details}
\label{implementation_details}

\textbf{Implementation setup.}
We use GPT-5.5 as the default LLM for GEMS and all cognitive baselines, ensuring that all methods are compared under the same LLM backbone. 
We use GPT-5.6 as the automatic evaluator and DeepSeek V4.1 Flash as the VLM. In addition, we use qwen3-embedding as the embedding model.
The default controller-bias coefficient is set to $\rho=0.5$, and the initial psychological state is $(3,3,3)$, corresponding to stress, depletion, and cognitive load, respectively. Action sample window $N$ is set to 3.
For the planning procedure $\mathcal{P}$, we use a rollout horizon of 30 steps.
For motion generation, we use Kimodo \citep{rempe2026kimodo} as the T2M model.
For ACTOR$^\dagger$, we follow the original ACTOR motion-clip sampling procedure, with the motion clips sampled from motions generated by Kimodo to ensure the same underlying motion-generation capability for comparison.

\textbf{Controlled-goal evaluation.}
We first evaluate all methods under a controlled-goal setting. All methods use the same persona, initial state, current time, and recent behavior history. Since different methods may vary in the range of activities they can generate or execute, we construct a shared daily-activity pool based on the semantic overlap among activities supported by different methods, including tasks such as preparing breakfast, cleaning the kitchen, and working from home. We then randomly sample 30 goals from this pool as the common test set for all methods.

For methods that support explicit task specification, including ACTOR, AGA, and X-VirtualHome, we directly provide the corresponding high-level goal. For desire-driven methods such as D2A and ASVO, we set the desire state associated with the target goal such that a semantically aligned behavior is triggered through the method's original decision-making mechanism. Once the target goal is explicitly provided or triggered through the desire state, we record the resulting behavior sequence until goal termination and evaluate only this goal-conditioned segment.

For each goal, each method is independently run five times, resulting in 150 behavior sequences per method. Since different human-like agent frameworks employ different cognitive modules and action representations, their generated behavior sequences may naturally differ in both action content and sequence length, even under the same goal and initialization. For each run, we use the behavior sequence produced by each method's own cognitive module as the target behavior description for its corresponding motion-generation module, and evaluate whether the system can faithfully realize its cognitive decisions as 3D behaviors. Specifically, SSR measures the fraction of successfully executed individual actions in the generated behavior sequence, GSR measures whether the entire generated behavior sequence is successfully completed, DTG measures the minimum wrist-to-goal distance during execution, and Qual. and Act. evaluate motion naturalness and consistency with the target behavior description, respectively.

This evaluation protocol is important because existing human-like agent frameworks differ substantially in their executable action spaces and action representations. For example, VirtualHome-based methods such as X-VirtualHome and AGA rely on predefined atomic action sets, while ACTOR relies on a library of pre-collected motion clips. These methods therefore generate, validate, and execute behaviors within the action spaces natively supported by their respective frameworks. Requiring all methods to execute an identical set of open-ended human-level action instructions would additionally penalize methods that cannot represent such actions due to their original design constraints. Conversely, restricting the evaluation to the intersection of predefined actions supported by all methods would artificially diminish the action-space extensibility that generative motion models are specifically intended to provide.

\textbf{Long-horizon evaluation.}
We additionally evaluate the methods under long-horizon daily simulation in Tab. \ref{tab:long_horizom__performance}. 
In this setting, we control only the persona and initial state across 
methods. 
For ACTOR and GEMS, which do not include a complete daily-schedule 
generation module, we construct a shared pool of high-level daily plans 
generated by the other methods and randomly sample a complete daily plan 
as input. 
ACTOR and GEMS then sequentially execute the activities specified in the 
sampled plan. 
Each simulation starts from the initial state of the day and continues 
until all scheduled daily activities are completed.

\subsection{Metric details.}
\label{metric_details}
For the cognitive module, we follow prior LLM-based evaluation 
protocols and report GPT-based scores on three dimensions. 
\textbf{Naturalness (Nat.)} measures the degree to which the generated behavior resembles real human behavior~\citep{wang2024d2a}.
A high Naturalness score indicates that the agent exhibits realistic human behavioral rhythms, rather than overly regular or mechanical patterns.
\textbf{Coherence (Coh.)} measures the logical and temporal consistency of the behavior sequence~\citep{wang2024d2a}.
A high Coherence score indicates that consecutive behaviors follow plausible causal relationships, with intermediate actions and contextual changes forming a convincing and coherent behavioral process.
\textbf{Personality Alignment (PA)} measures the consistency between the generated behavior and the given persona~\citep{li2025x}.
A high Personality Alignment score indicates that the agent's behavioral choices consistently reflect persona-specific preferences and tendencies, leading to distinguishable behavioral patterns across different personas.
\textbf{Mean} denotes the average of the three scores. 

\begin{figure}[h]
    \centering
    \includegraphics[width=0.85\linewidth]{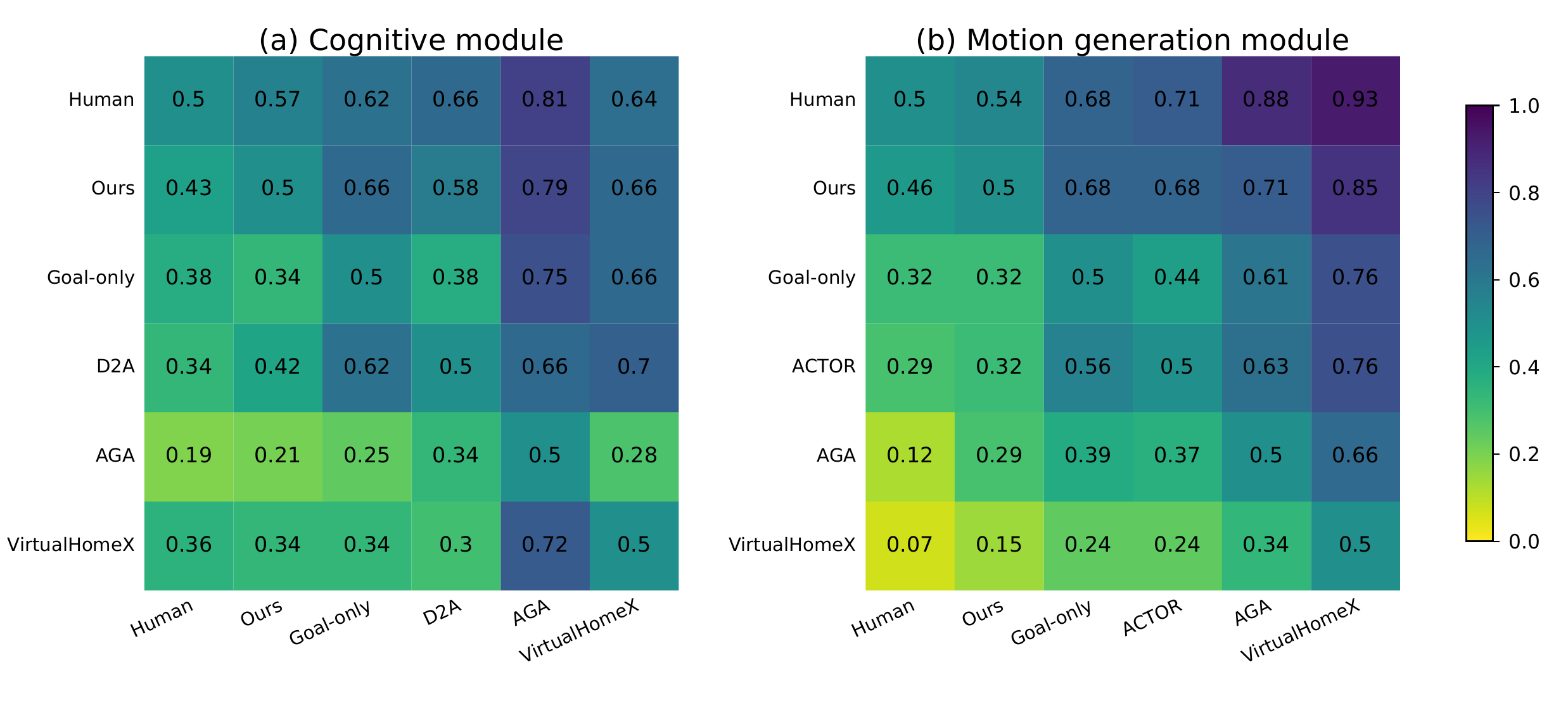}
    \caption{
        Human preference results from the subjective evaluation.
        The figure shows pairwise human-likeness comparisons for the Cognitive Behavior Simulation and 3D Motion Generation settings.
        Our full multiple controller system achieves preference scores close to real human behavior
        and consistently outperforms the Goal-only variant.
    }
    \label{fig:human_study_appendix}
\end{figure}

LLM-based evaluation protocols have been widely used in generative research as a scalable proxy for human user studies~\citep{cai2025diffusion}. The sequence-quality ranking obtained from the human subjective evaluation in Fig.~\ref{fig:human_study_appendix} is consistent with the ranking reported above, further supporting the validity of our evaluation protocol.

For the motion generation module, we evaluate both task execution and motion quality.
\textbf{Step Success Rate (SSR)} measures the percentage of steps for which the agent successfully achieves the corresponding step objective, based on a predefined contact-distance threshold. For example, a lying-down action is considered successful if both the hip and head joints of the human agent are within 30 cm of the target location~\citep{liang2025actor,hassan2023synthesizing}.
\textbf{Goal Success Rate (GSR)} measures the percentage of plans in which all steps are successfully executed~\citep{liang2025actor}.
\textbf{Distance to Goal (DTG)} measures the minimum wrist-to-goal distance during execution~\citep{diomataris2024wandr}.
For motion quality, given the rendered motion and its corresponding language description, human annotators rate each sample on a 1--5 scale in terms of (i) \textbf{Quality Score (Qual.)}, which evaluates the naturalness and plausibility of the generated motion in the given scene, and (ii) \textbf{Action Score (Act.)}, which evaluates how well the generated motion matches the action specified by the language description~\citep{wang2022humanise}. Higher scores indicate better motion quality and stronger consistency with the scene and language description.  Each sample is independently rated by three human annotators, and we report the average score across the three ratings.

\section{Human study}
\label{humanstudy}

\subsection{Human Evaluation}
\label{sec:human_evaluation}

\paragraph{Ethics and Participants.}
The study protocol was reviewed and approved by the ethics committee of our institution.
We recruited participants through Amazon Mechanical Turk (MTurk), an online crowdsourcing platform.
All participants provided informed consent before taking part in the study.
We obtained 53 valid responses for the Cognitive study and 41 valid responses for the 3D study.
All respondents reported an age of 18 years or older.
We additionally collected demographic information, including age, gender, ethnicity, and education level.
All responses were analyzed in de-identified form.

For the Cognitive study, the median reported age group was 25--29 years.
Among the 53 respondents, 27 identified as male, 14 as female, 6 as non-binary, and 6 preferred not to disclose their gender.
Regarding education, 24 respondents reported a postgraduate degree, 17 an undergraduate degree, 5 an A-Level, vocational, or technical qualification, and 3 a secondary-school or GCSE-level qualification; 4 preferred not to disclose their education level.
In terms of ethnicity, 26 respondents identified as White, 9 as Black, Black British, Caribbean, or African, 8 as Asian or Asian British, 6 as Mixed or Multiple ethnic groups, and 2 as another ethnic group; 2 preferred not to disclose their ethnicity.

For the 3D study, the median reported age group was also 25--29 years.
Among the 41 respondents, 24 identified as male, 13 as female, 3 as non-binary, and 1 preferred not to disclose their gender.
Regarding education, 17 respondents reported an undergraduate degree, 14 a postgraduate degree, 6 an A-Level, vocational, or technical qualification, and 1 a secondary-school or GCSE-level qualification; 3 preferred not to disclose their education level.
In terms of ethnicity, 18 respondents identified as White, 10 as Black, Black British, Caribbean, or African, 6 as Asian or Asian British, 6 as Mixed or Multiple ethnic groups, and 1 as another ethnic group.

\paragraph{Evaluation Protocol.}
We conduct two complementary human studies corresponding to the two stages of our evaluation.
The Cognitive setting evaluates the perceived human-likeness of behavioral sequences, while the 3D setting evaluates the perceived human-likeness of generated motion videos.
In the Cognitive setting, we compare GEMS, Goal-only, D2A, AGA, X-VirtualHome, and real human behavior.
In the 3D setting, we compare GEMS, Goal-only, ACTOR, AGA, X-VirtualHome, and real human behavior.
Goal-only is an ablated variant of GEMS that retains only the Goal-directed Controller.
Human references are selected from EgoLife~\citep{yang2025egolife} and TRUMANS~\citep{jiang2024scaling}.

For each setting, we perform all pairwise comparisons among the six behavior sources, resulting in 15 method pairs. For each method pair, we randomly sample 10 runs to construct the comparison pool. Method identities are hidden from participants. For each method pair, participants select the behavior they perceive as more human-like. Each valid response contains one judgment for each of the 15 method pairs.

For each behavior source, we select 10 behavioral samples for evaluation, resulting in 60 samples for each setting.
Each questionnaire presents samples from the six behavior sources and evaluates all pairwise combinations among them.
This produces $\binom{6}{2}=15$ pairwise comparisons per questionnaire.
The order and identities of the methods are hidden from participants, and participants are asked to select which of the two presented behaviors appears more human-like.
Thus, each valid response provides one judgment for each of the 15 method pairs.

For each method pair, we compute the pairwise preference rate as the fraction of participants preferring each method.
We further compute the overall preference rate of each method by aggregating its wins across its five pairwise comparisons with the other behavior sources.

\paragraph{Effect of Habitual Control.}
We assess the contribution of habitual control through a direct comparison between GEMS and Goal-only.
In the Cognitive setting, GEMS is preferred by 35 of 53 participants (66.0\%).
A two-sided exact binomial test against the null hypothesis of equal preference yields $p=0.027$.
In the 3D setting, GEMS is preferred by 28 of 41 participants (68.3\%), with $p=0.028$ under the same test.
These results show a consistent preference for GEMS over its Goal-only ablation in both behavioral sequences and generated 3D motions, supporting the contribution of habitual control to perceived human-likeness.

\paragraph{Comparison with Real Human Behavior.}
GEMS receives preference rates of 43.4\% and 46.3\% when directly compared with human references.

In the Cognitive setting, Human achieves an overall preference rate of 66.0\%, followed by GEMS at 62.6\%, D2A at 54.7\%, Goal-only at 50.2\%, X-VirtualHome at 41.1\%, and AGA at 25.3\%.
In the 3D setting, the corresponding rates are 74.6\% for Human, 67.8\% for GEMS, 51.2\% for ACTOR, 48.8\% for Goal-only, 36.6\% for AGA, and 21.0\% for X-VirtualHome.
Among the simulated methods evaluated in each setting, GEMS achieves the highest overall preference rate.
Together, these results indicate that GEMS can generate behaviors that are perceived as believable and human-like by participants in both behavioral-sequence and 3D-motion evaluations.

\paragraph{Consistency with Quantitative Evaluation.}
We further examine whether the human evaluation is consistent with the quantitative evaluation metrics.
In the Cognitive setting, the human-preference ordering of GEMS, D2A, X-VirtualHome, and AGA exactly matches the ordering obtained using the Mean score, yielding a Spearman rank correlation of $r_s=1.0$.

In the 3D setting, the human-preference ordering of GEMS, ACTOR, AGA, and X-VirtualHome is
$
\text{GEMS} > \text{ACTOR} > \text{AGA} > \text{X-VirtualHome},
$
and shows strong rank consistency with SSR, Qual., and Act., with $r_s=0.8$.
These results indicate that the quantitative metrics are broadly aligned with human judgments of behavioral and motion human-likeness.

\section{Case Study: Diverse Behavioral Patterns.}
\label{case_study2}

\begin{figure}[t]
    \centering
    \includegraphics[width=\linewidth]{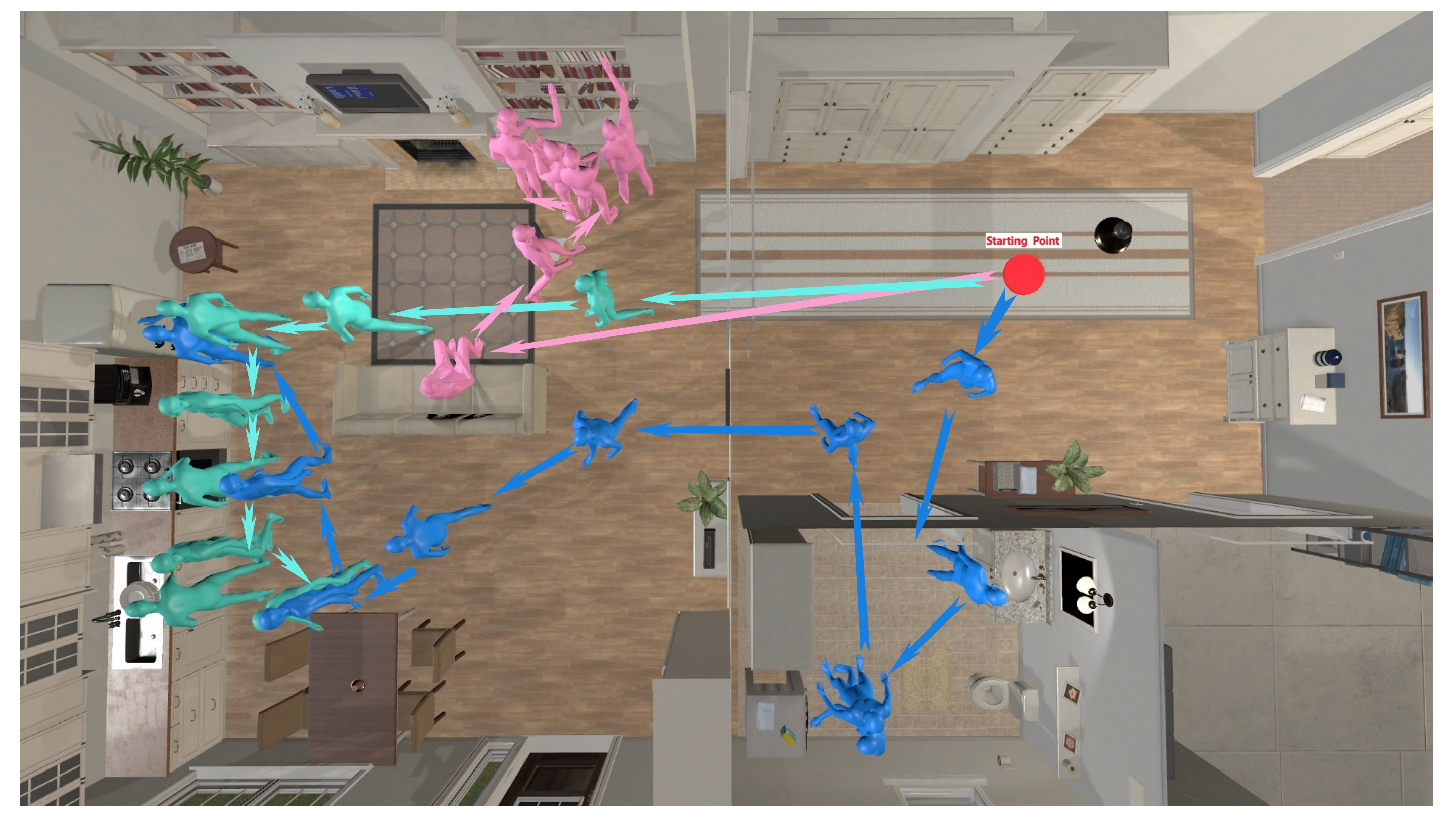}
    \caption{
Diverse behavioral trajectories generated under the same goal.
All three agents are assigned the goal of \textit{``deep clean the house''} and are initialized at the same spatial location, while differing in their personas, habit memories, and psychological states.
The spatial trajectories correspond to Adrian (colored green with $\rho=0.5$), Elena (colored blue with $\rho=0.6$), and Michael (colored pink with $\rho=0.4$).
For visual clarity, only the first six actions of each agent and their corresponding key action frames are shown.
The examples also illustrate cooperation and competition between habitual and goal-directed control.
}

    \label{fig:case_study2}
\end{figure}

We further conduct a case study to demonstrate how the proposed method generates diverse behavioral patterns under the same high-level goal. As shown in Fig.~\ref{fig:case_study2}, three agents are assigned the same goal, \textit{``deep clean the house''}, while having different personas, habitual associations, and internal psychological states. For visualization, we only show the first six actions of each agent and their corresponding execution trajectories.

\paragraph{Adrian (Green).}
Adrian has $\rho=0.5$, and his initial psychological state is set to 3. At the beginning of the task, he notices the refrigerator in his visual field and chooses to check it. This action is supported by both the habitual and goal-directed systems. Checking the refrigerator is relevant to the current cleaning goal and therefore receives a relatively high action value. Meanwhile, the cleaning context also activates one of Adrian's habitual behaviors. He then cleans the stove, which is mainly goal-directed. After completing this action, he adjusts his glasses as a habitual behavior. He next wipes the dishes. Since his hands become wet after this action, the resulting cue triggers another habit, causing him to casually fix his hair. He then returns to the cleaning task and wipes the table.

\paragraph{Elena (Blue).}
Elena has $\rho=0.6$, and her initial psychological state is also set to 3. She exhibits a different action sequence. She first cleans the sink and then hangs up the towel. Both actions directly contribute to the current cleaning goal. As her internal Dep. state increases, she stretches her shoulders, which is selected as a habitual action. She then wipes the table and finally checks the refrigerator, organizes it. Compared with Adrian, Elena shows a more continuous goal-directed trajectory during the first six actions, with fewer interruptions caused by habitual behaviors.

\paragraph{Michael (Pink).}
Michael has $\rho=0.4$, while his initial Dep. state is set to 5. This higher internal state leads to a substantially different trajectory. Although the same cleaning goal is active, he first goes to the sofa and sits down for a short rest. This habitual action is triggered by his adverse Dep. state. He then gets up and starts organizing the bookshelf. During this task, he notices a book and starts reading it, as the book acts as a contextual cue for one of his habitual behaviors. Reading further increases his cognitive load, which subsequently triggers another habitual action, touching his nose. Michael then resumes the task and continues organizing the bookshelf. Finally, the adverse Dep. state triggers another habitual response, and he sighs.

These examples show that GEMS can generate substantially different behavioral patterns. 
The case study also illustrates the cooperation and competition between the habitual and goal-directed systems. Adrian's first action provides an example of cooperation. Checking the refrigerator contributes to the cleaning goal and therefore receives support from the goal-directed controller. At the same time, the cleaning context activates the same action through Adrian's habitual controller. Both controllers therefore support the same behavior, increasing its likelihood of being selected.
Michael's first action provides an example of competition. Sitting on the sofa provides little immediate contribution to the cleaning goal and therefore has a relatively low goal-directed value. However, Michael's high Dep. state strongly activates the corresponding habitual response. The habitual controller consequently exerts a stronger influence on action selection at this moment, leading Michael to rest before continuing the task. Afterward, the goal-directed system drives him back to organizing the bookshelf.

\section{Additional discussions}
\label{discussions}

\subsection{Human-like LLM agent}

Human daily activity modeling supports a wide range of applications, from understanding everyday decision-making in psychology~\citep{garling1993psychological} and social interaction and structures in sociology~\citep{adler1987everyday}, to creating realistic virtual humans for metaverse and VR applications~\citep{schmidt2024frankenstein,fraser2024realistic}.
However, collecting real-world behavioral data is costly and raises privacy concerns. Recent advances in LLMs have therefore spurred growing interest in human behavior simulation.
The pioneering Generative Agents~\citep{park2023generative} introduced memory, reflection, and planning to simulate believable human behaviors. Building on this, Humanoid Agents~\citep{wang2023humanoid} incorporated desire-driven processes, while AFSPP~\citep{he2024afspp} introduced agent preferences and personalities to capture individual differences. More recent works model action selection from different psychological perspectives: CVA~\citep{zhang2026context} uses a human-data-trained Value Verifier, D2A~\citep{wang2024d2a} adopts desire-driven behavior based on the Theory of Needs, and ASVO~\citep{lin2026asvo} incorporates Social Value Orientation for multi-agent interactions.
For 3D realization, X-VirtualHome~\citep{leng2026agentsense} generates persona-conditioned routines with LLM-based planning, while AGA~\citep{yu2024affordable} grounds high-level instructions into VirtualHome's predefined action space. X's Day~\citep{li2025x} models personality-driven long-term behaviors, Visually-grounded Humanoid Agents~\citep{ye2026visually} grounds LLM-driven behaviors through visual perception and spatial planning, and Ella~\citep{zhang2025ella} introduces long-term multimodal memory for lifelong embodied behavior. Most of these frameworks realize high-level behaviors through predefined atomic actions. In contrast, ACTOR~\citep{liang2025actor} matches planned behaviors with pre-collected motion clips and uses Transformer-based motion completion to connect them, which can introduce spatial misalignment and skeleton drift during transitions. Note that existing methods are predominantly driven by explicit plans, desires, or events, resulting in a goal-directed behavior generation paradigm. Habitual behaviors, however, are often cue-triggered, non-optimal, and highly individualized, accounting for a major portion of everyday human behavior. Modeling habits can therefore enable more natural and personalized agents and benefit human behavior understanding and recognition.

\subsection{Human behavior control study}

Goal-directed and habitual control have long been viewed as two ends of a spectrum of human behavioral control~\citep{dolan2013goals}. The same action may arise from goal-directed control, habitual control, or their joint influence~\citep{miller2019habits,wood2022habits}, making the underlying source of control difficult to infer from overt behavior alone. Moreover, habits are often executed automatically with limited deliberation, while their underlying triggers may escape conscious awareness, leading their influence on behavior to be underestimated~\citep{wood2016psychology}. Experimental studies commonly distinguish the two forms of control using outcome devaluation, which tests sensitivity to current outcome value~\citep{dolan2013goals}. 
In cognitive neuroscience, human behavioral control is often modeled through reinforcement learning: goal-directed control is associated with model-based planning, whereas habitual control is associated with model-free learning~\citep{daw2005uncertainty,dolan2013goals,wood2016psychology}. Early dual-system models emphasized competition between the two controllers, while later work introduced arbitration mechanisms that dynamically balance their influence based on relative reliability~\citep{daw2005uncertainty,lee2014neural}. Related cognitive architectures exploit proceduralized responses for efficient action selection: ACT-R enables perceptual cues to trigger productions after practice~\citep{taatgen2008acquisition}, Soar represents procedural knowledge with production rules~\citep{laird1987soar}, and CoALA extends these principles to language agents~\citep{sumers2024cognitive}.
More recent accounts question the direct equivalence between habits and model-free RL, instead linking habitual behavior to habit strength shaped through repetition~\citep{miller2019habits}. However, existing studies mainly focus on habit formation and control allocation in constrained tasks, leaving contextual habit triggering and state-dependent shifts in realistic environments underexplored~\citep{wood2016psychology,neal2013people}.
In contrast, we study multiple behavioral control system for human-like daily behavior simulation in realistic 3D environments.

\subsection{Underestimation of Habitual Influences}
\label{challengeofhabits}

Habitual influences on behavior may be systematically underestimated because people have limited introspective access to the mechanisms that trigger habits. \citet{wood2016psychology} argue that people are often aware of performing a habitual response, yet remain largely unaware of the contextual cues that activate it. As a result, explanations for habitual behavior can become largely post hoc: people observe what they repeatedly do and infer the goals or intentions that might have caused it. Indeed, people with stronger habits reported greater certainty about their intentions and perceived their behavior as more goal guided, even though intentions were actually poorer predictors of strongly habitual behavior. This tendency may be strengthened by the intuitive appeal of goal-based explanations. \citet{wood2022habits} note that people spontaneously infer goals behind behavior and generally favor goal explanations over alternative causes. Their coffee-drinking study provides a clear example. Although habit and fatigue had comparably sized effects on actual coffee consumption, participants attributed their coffee drinking much more strongly to fatigue than to habit. This underestimation of habit persisted even when participants were financially incentivized to make accurate judgments. Together, these findings suggest that people may experience habitual behavior through a goal-based narrative: because the automatic context–response process is relatively inaccessible to introspection, people infer a plausible motive after the behavior occurs. Consequently, they may overestimate the role of goal pursuit while underestimating the contribution of habit to everyday action.

\subsection{Overall discussion}
\label{more_discussion}
This work aims to develop a human control mechanism inspired framework for generating more believable human-like behavior. Believability relies on sustaining an “illusion of life,” in which an agent exhibits stable, natural, and individually distinctive behavioral patterns \citep{park2023generative,bates1994role}. To support this goal, we draw on research on goal-directed and habitual behavior to derive a multiple control system design.

Human action control is commonly described as involving interacting goal-directed and habitual systems. These systems rely on partially distinct psychological and neural mechanisms while jointly shaping behavior. Goal-directed control is more strongly associated with prefrontal and anterior striatal regions, whereas habitual control is linked to posterior striatal regions, particularly the posterior putamen \citep{daw2005uncertainty}.
Computationally, goal-directed control is often mapped onto model-based control, while habitual control is mapped onto model-free control. Model-based control uses an internal model of the environment to predict future states and outcomes, whereas model-free control relies on cached action values learned through experience. \citet{daw2005uncertainty} further proposed an arbitration mechanism that dynamically allocates control according to the relative reliability of the two systems. This motivates our architecture consisting of a goal controller, a habit controller, and an arbiter.
For the habit controller, we further follow value-free account of habit. Standard model-free reinforcement learning still represents actions through expected future reward. \citet{miller2019habits} instead argue that habits can arise from repeated actions without explicitly encoding their outcome values. Accordingly, our framework represents habitual tendencies through habit strength rather than model-free cached value.

Classical RL models nevertheless remain difficult to apply directly to open-world embodied environments, where states and actions are not fixed and are often represented through natural language. We therefore extend the multiple system architecture to an LLM-based agent and incorporate four additional principles from the psychology of habit:

\textit{Context-driven habit retrieval.} Habits can be rapidly triggered by familiar contextual cues and can be activated independently of current goals \citep{wood2016psychology}. We therefore explicitly retrieve habitual actions from the current context.
\textit{Asymmetric interaction between habits and goals.} Habitual responses are activated quickly and can narrow the space of candidate actions, whereas goal-directed control requires additional evaluation of possible outcomes \citep{wood2016psychology}. We therefore allow the habit controller to propose candidate actions that are subsequently evaluated by the goal controller.
\textit{Dynamic modulation by psychological state.} Time pressure, distraction, and cognitive load can weaken goal-directed control and increase reliance on habits. Habit slips illustrate this imbalance, as previously learned responses may remain active even after current goals have changed \citep{wood2022habits}. We abstract these factors into a current psychological state, which dynamically modulates the relative influence of the two controllers.
\textit{Stable individual differences in control bias.} Individuals differ in their tendency to rely on goal-directed versus habitual control. We represent this stable variation using an individual control bias \(\rho\), allowing different agents to exhibit distinct control tendencies under the same environmental and psychological conditions \citep{lee2014neural}.

In summary, this work aims to bring a multiple control perspective to the human-like agent community. We believe we provide useful insights for future research on believable and human-like agent behavior.

\section{Habit controller}
\label{habitcontroller}

\subsection{Habit Dataset Construction}
\label{app:habit_dataset}

We construct the Habit Dataset to address the difficulty of distinguishing habitual behavior from goal-directed behavior based solely on observation, as discussed in \ref{challengeofhabits}. People often have limited awareness of the automatic mechanisms that trigger habitual behavior and tend to attribute repeated actions to goals or intentions. As a result, inferring habits directly from observed daily behavior can lead to attribution ambiguity. To reduce this ambiguity, we collect behavioral instances from empirical sources that explicitly investigate habitual behavior, including habit scales, behavioral studies, and publicly available datasets, and further extract their associated context cues. This design provides clearer empirical evidence that the collected behaviors are habit-related and improves the controllability and reliability of habit influence in the subsequent generation of daily human behavior.

\paragraph{Context Cue Representation.}
Potentially, any internal or external event may trigger habitual behavior~\citep{gardner2016habitual}.
Since the full space of possible cues cannot be exhaustively represented, we consider four cue dimensions that are both empirically supported and available in the agent's current context: visual episodes~\citep{hommel1998event}, psychological states~\citep{waszak2009episodic}, events~\citep{qiu2023influence}, and active goals~\citep{schneider2009selecting}.
Each cue is represented in natural language and assigned to one of these dimensions.
Although a habitual response may depend on multiple contextual factors, we use one primary cue dimension for each habit instance to obtain a simple and interpretable representation.

\paragraph{Habit Dataset Overview.}
We construct a dataset of daily habitual behaviors and their associated context cues from existing empirical evidence, including habit scales, behavioral studies, and publicly available habit-related datasets~\citep{ersche2017creature,moors2006automaticity,georgiev2022development,li2024habitaction,pang2022individual,azrin1973habit,wood2016psychology}.
Each instance is represented as a cue--action pair
$\langle q_i^{cue}, a_i^H \rangle$.
After extraction, standardization, deduplication, and manual validation, the resulting dataset contains 571 habitual behavior instances.
The following paragraphs describe the construction procedure in detail.

\textbf{Data Collection.}
We collect raw habitual behavior data from existing sources of empirical
evidence, including habit scales, empirical psychological studies, and publicly
available habitual behavior datasets. All included sources explicitly describe
or study habitual behaviors in daily life and provide contextual information
about the cues under which these habits are expressed or triggered.

The collected behaviors mainly cover two categories considered in our setting:
object-interaction habits and self-touch habits. Object-interaction habits
include, for example, eating a visible snack, drinking a beverage under a
particular psychological state, or checking a watch while waiting. Self-touch
habits include behaviors such as scratching one's head under high cognitive
load or wiping one's mouth after drinking.

Different sources provide information at different levels of granularity,
including habitual actions, behavioral contexts, experimental descriptions,
and video annotations. We retain the original behavior descriptions,
contextual information, and source information for traceability.

\textbf{LLM-based Information Extraction.}
The raw source materials come from multiple sources and are presented in
heterogeneous forms, including paper text, tables, and dataset annotations.
We therefore use an LLM to extract the habitual action, context cue, and
corresponding source evidence, and convert them into a unified structured
representation.

\medskip
\noindent\fbox{
\begin{minipage}{0.94\linewidth}
\textbf{Example -- Raw extraction}

\medskip
\texttt{Source text: ``anxiety increases the frequency of self-touch''}\\
\texttt{Context cue: feeling anxious}\\
\texttt{Habitual action: self-touch}
\end{minipage}
}
\medskip

\textbf{Standardization and Cleaning.}
We map each extracted cue to one of the four predefined cue dimensions used in our framework. These cue dimensions have been extensively studied as contextual factors associated with habitual or stimulus--response behavior~\citep{gardner2016habitual,wood2016psychology}, including \textbf{visual episodes}~\citep{hommel1998event}, such as observing that a glass is empty and being cued to refill it; \textbf{psychological states}~\citep{waszak2009episodic}, such as experiencing high cognitive load and being cued to touch one's head; \textbf{events}~\citep{qiu2023influence}, such as finishing a drink and being cued to wipe one's mouth; and \textbf{active goals}~\citep{schneider2009selecting}, such as having the current goal of cleaning the kitchen and being cued to take out the trash. Both context cues and habitual actions are then normalized into natural-language descriptions. For instances with missing cues, annotators inspect the original paper, video content, dataset description, and related evidence to determine whether a supported triggering cue can be identified. Instances without reliable cue evidence are removed. We also remove instances whose cues cannot be reliably mapped to one of the four predefined cue dimensions. We further apply missing-field filtering, language normalization, and semantic deduplication to obtain standardized context cue--habitual action pairs.

Our primary objective is to study whether introducing habitual control improves
the human-likeness of generated behavior. We therefore adopt a simplified
representation in which each habit instance retains one primary cue dimension.
Despite this simplification, the resulting framework produces clear
improvements in human-likeness in our experiments, supporting the effectiveness
of the proposed framework. Modeling compound context cues across multiple
dimensions is left to future work for finer-grained habit triggering.

\medskip
\noindent\fbox{
\begin{minipage}{0.94\linewidth}
\textbf{Example -- Standardized pair}

\medskip
\texttt{Cue dimension: psychological state}\\
\texttt{Context cue: ``The person feels anxious.''}\\
\texttt{Habitual action: ``Engage in self-touch.''}
\end{minipage}
}
\medskip

\textbf{Quality Validation.}
Because the processed dataset is moderate in size, we manually validate all
processed instances and remove inappropriate or insufficiently supported
instances. The inspection considers three aspects: (1) whether the habitual
action is supported by the original habit-related source; (2) whether the
extracted cue--action association is consistent with the source evidence; and
(3) whether the context cue is assigned to the correct cue dimension.

After this process, we obtain 571 valid context cue--habitual action instances
and use them to construct the global habit dataset $ \mathcal{D}^H $.

\medskip
\noindent\fbox{
\begin{minipage}{0.94\linewidth}
\textbf{Data examples}

\medskip
\texttt{ContextCue(visual\_episode = ``A glass of water is visible.'')}\\
\texttt{habitual\_action = ``Drink water.''}

\medskip
\texttt{ContextCue(psychological\_state = ``The person feels anxious.'')}\\
\texttt{habitual\_action = ``Engage in self-touch.''}
\end{minipage}
}

\subsection{Personalized Habit Memory Initialization}
\label{app:habit_initialization}

Habit memory is inherently dynamic. A habit may gradually weaken or even be forgotten due to interference from external factors, while others may be further reinforced through long-term repetition. However, this work focuses on fine-grained human behavior simulation over relatively short time horizons, typically within a single day. Therefore, habit memory can be reasonably assumed to remain stable over the simulated time period.

With a convincing Habit Dataset, we can provide agents with daily habitual behaviors that are observed in real human life. This reduces the risk that our framework confuses habitual behavior with goal-directed behavior during generation. Specifically, we sample a subset of habitual behaviors from the Habit Dataset for each agent and assign a habit strength to each behavior.

It is important to clarify that habits are typically acquired gradually through repeated practice over an extended period of time. Habit formation is therefore a long-term learning process, and this process has already been extensively studied in existing computational frameworks \citep{miller2019habits}. Our focus is different. We investigate the role of habitual behavior as a behavioral driver in human-like behavior generation, rather than re-modeling the full process of habit formation. Therefore, we use an LLM to assign habitual behaviors and their corresponding strengths to each agent as an approximation of an already established habit state. This design is sufficient for studying whether incorporating habitual behavior can improve the believability and human-likeness of generated behavior. The effectiveness of this design is further supported by our experimental results.

\textbf{Persona Construction.}
To model long-term individual differences, we construct a structured persona
for each agent, including demographic attributes, personality, occupation, and
long-term background. We also initialize the agent's momentary psychological
states, including stress, depletion, and cognitive load, which are subsequently
used for behavioral control.

\medskip
\noindent\fbox{
\begin{minipage}{0.94\linewidth}
\textbf{Example persona}

\medskip
\textbf{Michael Anderson (Michael), 37, male.}

\textbf{Personality:}
Gentle, responsible, cautious, pragmatic, stability-oriented,
conflict-avoidant, and mildly anxious; sensitive to risk and criticism.

\textbf{Occupation:}
Project Coordinator in the City Planning Department of Portland, Oregon,
responsible for community communication, resident feedback, and
cross-department coordination.

\textbf{Initial Internal State:}
Stress 3/10; Depletion 5/10; Cognitive Load 3/10.

\textbf{Background:}
Raised in a middle-class family near Portland and trained in geography and
urban studies. He values family security, health, savings, and long-term
stability, and generally prefers reliable, low-risk choices.
\end{minipage}
}
\medskip

\textbf{Persona-conditioned Habit Sampling.}


Given an agent persona, we construct its personalized habit memory from the
global habit dataset $\mathcal{D}^H$. Since $\mathcal{D}^H$ contains 571
candidate context cue--habitual action pairs, we directly provide the agent's
long-term persona together with all candidate pairs to an LLM. The LLM selects
30 habit instances for each agent. All agents use the same habit memory
capacity.

We use the following prompt:

\medskip
\noindent\fbox{
\begin{minipage}{0.94\linewidth}
You will be given a persona profile and a set of candidate
context cue--habitual action pairs. Select the habits that are most appropriate
for the long-term characteristics of this persona.

When selecting habits, consider the following three aspects:

\textbf{1. Consistency:}
The selected habits should be consistent with the persona's personality,
experience, lifestyle, and long-term behavioral tendencies, and should not
clearly conflict with the persona description.

\textbf{2. Plausibility:}
The selected behaviors should be habits that the persona could plausibly have
formed through repeated experience over time. Do not assume that a persona has
a habit simply because a context cue resembles its current state.

\textbf{3. Diversity:}
The selected habits should cover different types of daily behaviors and
context cues, while avoiding semantically redundant behaviors or excessive
concentration in the same type of situation.

Select the most suitable habits from the candidate set according to the above
criteria. Return only the IDs of the selected candidates. Do not modify the
context cue or habitual action, and do not generate habits that are not present
in the candidate set.
\end{minipage}
}
\medskip

\textbf{Habit Strength Assignment.}
For the 30 selected habit instances, we further assign an agent-specific habit
strength $H(a_i^H)\in[0,1]$ to each habitual action $a_i^H$. Habit strength is
a dimensionless normalized scalar that represents the relative strength of the
context--response association. We condition this estimation on the agent's
long-term personality, experience, and living environment.

We use the following prompt:

\medskip
\noindent\fbox{
\begin{minipage}{0.94\linewidth}
Given a persona's personality, long-term experience, and living environment,
together with a set of context cue--habitual action pairs, estimate a habit
strength $H(a_i^H)$ between 0 and 1 for each habit.

Habit strength represents the relative strength of the context--response
association. Consider the following factors:

\textbf{1. Repetition:}
How likely the behavior is to have been repeatedly performed in the persona's
long-term daily life.

\textbf{2. Context Stability:}
How consistently and repeatedly the corresponding context cue is likely to
occur in the persona's life.

\textbf{3. Long-term Consistency:}
How consistent the habit is with the persona's personality, experience,
occupation, lifestyle, and long-term behavioral tendencies.

Use the following scale:

0.0--0.4: very weak association;\\
0.4--0.7: moderate association;\\
0.7--0.9: stable habit;\\
0.9--1.0: strong habit.
\end{minipage}
}
\medskip

After habit sampling and strength assignment, we obtain the personalized habit
memory $ \widehat{\mathcal{D}}^H $ for each agent.

\medskip
\noindent\fbox{
\begin{minipage}{0.94\linewidth}
\textbf{Habit memory examples}

\medskip
\texttt{HabitMemoryItem(}\\
\texttt{\quad context\_cue = ContextCue(}\\
\texttt{\qquad visual\_episode = ``A glass of water is visible.''}\\
\texttt{\quad ),}\\
\texttt{\quad habitual\_action = ``Drink water.'',}\\
\texttt{\quad habit\_strength = 0.80}\\
\texttt{)}

\medskip
\texttt{HabitMemoryItem(}\\
\texttt{\quad context\_cue = ContextCue(}\\
\texttt{\qquad psychological\_state = ``The person feels anxious.''}\\
\texttt{\quad ),}\\
\texttt{\quad habitual\_action = ``Engage in self-touch.'',}\\
\texttt{\quad habit\_strength = 0.92}\\
\texttt{)}
\end{minipage}
}

\begin{table}[t]
\centering
\caption{Comparison between LLM-based and human-annotated habit initialization for Alice. Mean denotes the average of Nat., Coh., and PA.}
\label{tab:habit_init}
\begin{tabular}{lcccc}
\toprule
Habit initialization & Nat. $\uparrow$ & Coh. $\uparrow$ & PA $\uparrow$ & Mean $\uparrow$ \\
\midrule
LLM-based       & 87.5 & 88.2 & 85.8 & 87.2 \\
Human-annotated & 86.3 & 89.1 & 87.7 & 87.7 \\
\bottomrule
\end{tabular}
\end{table}

To further validate the LLM-based habit initialization scheme, we additionally construct a manually designed agent, Alice. Three volunteers jointly determine Alice's habit memory and the corresponding habit strengths based on her persona, which we use as the Human-annotated initialization. We compare this initialization with the LLM-based scheme adopted in our framework. As shown in Tab.~\ref{tab:habit_init}, the two initialization strategies achieve comparable performance in Naturalness, Coherence, and Personality Alignment. These results indicate that the LLM-based habit initialization scheme is a reasonable approach for personalized habit assignment and habit-strength estimation.

\subsection{Habit Retrieval Evaluation Set}
\label{app:habit_retrieval}

To evaluate the cue retrieval mechanism of the Habitual Controller, we construct an agent-specific retrieval evaluation set based on each personalized habit memory $\hat{D}^{H}$. Each agent contains 30 habit memories. Each memory consists of a context cue $q_i^{\mathrm{cue}}$, a habitual action $a_i^{H}$, and its habit strength $H(a_i^{H})$. Since different agents have different personalized habit memories, we construct the retrieval evaluation set independently for each agent.

For each habit, we construct four types of retrieval cases. Each case has one ground-truth retrieval label, which can be either a target habitual action or $\emptyset$, indicating that no habitual action should be retrieved. \textit{Direct Match} provides a context that directly matches the cue of the target habit. It tests whether the retrieval algorithm can identify the correct habit under a clear triggering context. \textit{Boundary Reversal} changes a key contextual attribute, such as object state, event timing, or subject identity. This creates a more confusing context and tests whether the algorithm can use the key contextual information to identify the target habit. \textit{Partial Transfer} preserves part of the semantic or categorical similarity while changing the location, object, state, intensity, or event source. It tests whether the algorithm can distinguish the target habit from other similar habits under partial contextual changes. \textit{Hard Negative} introduces strong distractors that are semantically similar to existing habit cues or contain misleading contextual information, while not satisfying the actual triggering conditions of any habit. Examples include objects shown only in photographs, descriptions of psychological states, or objects with similar but different states. These cases test whether the retrieval algorithm can correctly reject misleading cues and avoid triggering any habitual action under highly confusing contexts.

For each retrieval case, the ground-truth label is the corresponding target habitual action, or $\emptyset$ for a hard-negative case. A retrieval trial is considered correct if its output matches the ground-truth label; otherwise, it is considered incorrect. Therefore, the evaluation directly measures whether the retrieval algorithm can identify the correct habitual behavior or correctly return no habitual action under different types of contextual variation and interference.

This process produces $30 \times 4 = 120$ base retrieval cases for each agent. Among these 120 base retrieval cases, we deliberately construct a subset of challenging cases in which multiple candidate habits have high cue compatibility with the current context. Some cases are also designed such that cue compatibility and habit strength favor different candidate habits. These challenging constructions test whether the retrieval algorithm is affected by strong distractors, high habit strength, or superficial semantic similarity.

To evaluate robustness to linguistic variation, we generate four linguistic variants for each base case. These variants modify the wording, temporal descriptions, and irrelevant background information while keeping the same ground-truth retrieval label. Each agent therefore contains $120 \times 4 = 480$ retrieval cases. We independently construct the evaluation sets for three agents, resulting in 1,440 retrieval cases in total. We randomly inspect 10\% of the generated cases to verify the context descriptions, case types, and ground-truth retrieval labels.

For evaluation, we split the 120 base cases of each agent into validation and test sets using a 60\%/40\% split. All four linguistic variants derived from the same base case are assigned to the same split, resulting in 288 validation cases and 192 test cases per agent. The validation set is used to select the retrieval threshold $\kappa$, while the test set is used only for reporting the final retrieval performance.

\subsection{Retrieval Algorithm Evaluation}
\label{app:retrieval_algorithm}

Habit retrieval and selection are mainly influenced by two factors: the compatibility between the current context and the stored context cue, and the strength of the habit itself. We use cue similarity to characterize context--cue compatibility, which measures how well the current context matches the stored cue of a habit. Habit strength characterizes the strength of the stimulus--response (S--R) association. A stronger S--R association indicates a higher tendency for the corresponding habitual response to occur and guide behavior. Based on these two factors, we compare three retrieval strategies to examine their different roles in habit retrieval and selection.

We first introduce \textit{Similarity-only} as a baseline. This method performs both retrieval and selection solely according to cue similarity, allowing us to isolate the contribution of habit strength. We then consider a direct combination of the two factors, denoted as \textit{Similarity $\times$ Habit Strength}, which multiplies cue similarity by habit strength before retrieval. In this formulation, cue compatibility and habit strength jointly affect habit activation from the beginning. However, a high habit strength may compensate for relatively weak cue compatibility, while a low habit strength may prevent a well-matched cue from being retrieved.

Motivated by the observation that habitual responses are triggered by recurring context cues, while stronger habits are more likely to guide behavior~\citep{wood2016psychology}, we further introduce \textit{Two-stage Cue Retrieval}. The first stage uses only cue similarity to identify habits that are sufficiently compatible with the current context and constructs the set of activated habits. The second stage incorporates habit strength to weight and select among these activated habits. This design separates context--cue matching from the effect of habit strength. Habit strength therefore does not interfere with the initial cue matching, while stronger habits remain more likely to be selected after activation.

We next describe the three retrieval strategies in detail. Given the current context $s_t$ and the personalized habit memory $\hat{D}^{H}$, the retrieval algorithm compares the current context with the stored context cues and returns either a corresponding habitual action $a_{i,t}^H$ or no habitual action if no habit is retrieved.

We consider four types of context cues for habit triggering: visual episodes, psychological states, events, and active goals. Each stored context cue $q_i^{\mathrm{cue}}$ belongs to one of these cue dimensions. For habit $i$, we use $c_{i,t}$ to denote the information extracted from the current context $s_t$ that belongs to the same cue dimension as $q_i^{\mathrm{cue}}$. For example, if $q_i^{\mathrm{cue}}$ corresponds to a psychological state, $c_{i,t}$ represents the psychological-state information in the current context.

We encode $c_{i,t}$ and $q_i^{\mathrm{cue}}$ into the same embedding space, obtaining vector representations $\mathbf{c}_{i,t}$ and $\mathbf{q}_i^{\mathrm{cue}}$. Their cue similarity is measured using cosine similarity, i.e.,
$
C_i =
\frac{
\mathbf{c}_{i,t}^{\top}\mathbf{q}_i^{\mathrm{cue}}
}{
\|\mathbf{c}_{i,t}\|_2
\|\mathbf{q}_i^{\mathrm{cue}}\|_2
}.
$
A larger $C_i$ indicates higher compatibility between the current context and the stored cue of habit $i$. All three retrieval strategies use $C_i$, but differ in whether and when habit strength $H(a_i^H)$ contributes to retrieval and selection.

\paragraph{Similarity-only.}
This baseline relies solely on cue similarity for both retrieval and selection. We retain all habits whose cue similarity is at least the threshold $\kappa$, forming the activated set
$
\mathcal{E}_t = \{i \mid C_i \geq \kappa\}.
$
If $\mathcal{E}_t=\emptyset$, no habitual action is retrieved. For each habit $i\in\mathcal{E}_t$, we compute its response score as
$
p_i = \frac{C_i-\kappa}{1-\kappa}.
$
The response scores are normalized into selection probabilities, i.e.,
$
P(a_{i,t}^H \mid s_t,\mathcal{E}_t)
=
\frac{p_i}
{\sum_{j\in\mathcal{E}_t}p_j}.
$
A habitual action is then sampled according to this distribution. This method does not use habit strength during either retrieval or selection.

\paragraph{Similarity $\times$ Habit Strength.}
This baseline directly combines cue similarity and habit strength before retrieval. For each habit $i$, we first compute the joint score
$
S_i = C_i H(a_i^H).
$
We then construct the activated set as
$
\mathcal{E}_t = \{i \mid S_i \geq \kappa\}.
$
If $\mathcal{E}_t=\emptyset$, no habitual action is retrieved. For each $i\in\mathcal{E}_t$, the response score is
$
p_i = \frac{S_i-\kappa}{1-\kappa},
$
which is normalized as
$
P(a_{i,t}^H \mid s_t,\mathcal{E}_t)
=
\frac{p_i}
{\sum_{j\in\mathcal{E}_t}p_j}.
$
A habitual action is then sampled according to this distribution. In this strategy, habit strength directly contributes to the retrieval score and therefore also affects whether a habit enters the activated set.

\paragraph{Two-stage Cue Retrieval (Ours).}
Our method separates cue matching from the effect of habit strength. In the first stage, we use only cue similarity to construct the activated set:
$
\mathcal{E}_t = \{i \mid C_i \geq \kappa\}.
$
If $\mathcal{E}_t=\emptyset$, no habitual action is retrieved. In the second stage, habit strength is introduced only for habits that have already entered $\mathcal{E}_t$. For each $i\in\mathcal{E}_t$, we compute
$
p_i =
\frac{C_i-\kappa}{1-\kappa}
H(a_i^H),
$
and normalize the response scores as
$
P(a_{i,t}^H \mid s_t,\mathcal{E}_t)
=
\frac{p_i}
{\sum_{j\in\mathcal{E}_t}p_j}.
$
A habitual action is then sampled according to this distribution. This two-stage design prevents habit strength from interfering with context--cue matching, while still allowing habit strength to influence the subsequent action selection. As a result, habits with stronger S--R associations have a higher tendency to be selected once they have been activated.

\paragraph{Retrieval Accuracy.}

We evaluate the three retrieval strategies using the agent-specific validation and test sets described above. Each retrieval case has one ground-truth label, which can be either a habitual action or no habitual action. For agent $j$, let $a_{j,n}^{H,*}$ denote the ground-truth label of the $n$-th retrieval case, where $a_{j,n}^{H,*}=\emptyset$ if no habitual action should be retrieved. Since habitual action selection is stochastic, for each retrieval strategy and each evaluated threshold $\kappa$, we independently sample the retrieval outcome 10 times for each case. Let $\hat{a}_{j,n,r}^{H}(\kappa)$ denote the output returned in the $r$-th retrieval trial, where $r=1,\ldots,10$. If the algorithm retrieves no habitual action in a trial, we set $\hat{a}_{j,n,r}^{H}(\kappa)=\emptyset$. Each retrieval trial is counted as correct when its output matches the ground-truth label; otherwise, it is counted as incorrect. Retrieval Accuracy is computed by averaging correctness over the 10 trials for each case and then over all cases.

For both the validation and test sets, Retrieval Accuracy is computed in the same manner. Each agent contains 288 validation cases and 192 held-out test cases. For each retrieval strategy, we first evaluate Retrieval Accuracy under different values of $\kappa$ ($\kappa < 1$) on the validation set and select the threshold that gives the highest validation accuracy for each agent. Since each agent has a different personalized habit memory, the selected retrieval threshold can differ across agents. After selecting the threshold, we keep it fixed and evaluate the corresponding retrieval strategy on the held-out test set. The Retrieval Acc. reported in the main paper is the average of the test-set Retrieval Accuracy across the three agents at their validation-selected thresholds.

Tab.~\ref{tab:retrieval_threshold_test} reports the best validation accuracy, the corresponding selected threshold $\kappa^*$, and the held-out test accuracy for each agent and retrieval strategy. We search $\kappa$ with an interval of $0.05$ on the validation set and then fix the selected $\kappa^*$ for evaluation on the test set. Across the three agents, our Two-stage Cue Retrieval selects consistently lower thresholds than the two baselines. This pattern is consistent with our two-stage design, where $\kappa$ is applied only to cue similarity during habit activation, while habit strength is introduced only after a habit has been activated. This separation prevents habit strength from interfering with the initial cue-matching stage and allows the threshold to focus on whether the contextual cue is sufficiently compatible. With the validation-selected thresholds fixed, our method achieves an average test accuracy of $84.7\%$, compared with $71.4\%$ for Sim.~$\times$~Habit Strength and $73.3\%$ for Similarity-only. The final results are reported in Tab.~\ref{tab:ablation_e}.

\begin{table*}[t]
\centering
\caption{Validation-selected retrieval thresholds and held-out test performance.
For each agent and each retrieval strategy, $\kappa$ is selected on the
validation set and fixed for test evaluation.}
\label{tab:retrieval_threshold_test}

\small
\setlength{\tabcolsep}{5pt}
\renewcommand{\arraystretch}{1.05}

\begin{tabular}{lccccccccc}
\toprule
& \multicolumn{3}{c}{Ours}
& \multicolumn{3}{c}{Sim.$\times$HS}
& \multicolumn{3}{c}{Sim.-only} \\
\cmidrule(lr){2-4}
\cmidrule(lr){5-7}
\cmidrule(lr){8-10}

Agent
& Val. & $\kappa^*$ & Test
& Val. & $\kappa^*$ & Test
& Val. & $\kappa^*$ & Test \\
\midrule

Michael
& \textbf{82.8} & 0.55 & \textbf{81.2}
& 71.0 & 0.80 & 71.6
& 73.0 & 0.85 & 72.1 \\

Adrian
& \textbf{85.4} & 0.55 & \textbf{82.1}
& 68.7 & 0.70 & 66.9
& 71.1 & 0.90 & 70.4 \\

Elena
& \textbf{93.1} & 0.45 & \textbf{90.7}
& 77.1 & 0.75 & 75.8
& 78.6 & 0.85 & 77.3 \\

\midrule

\textbf{Average}
& \textbf{87.1} & & \textbf{84.7}
& 72.3 & & 71.4
& 74.2 & & 73.3 \\

\bottomrule
\end{tabular}
\end{table*}

\section{Design of the Psychological state function in the arbiter}
\label{gate}

The effects of momentary psychological states on the balance between goal-directed and habitual control are complex. Existing computational models have mainly focused on how factors such as controller reliability, uncertainty, and task complexity influence the allocation of behavioral control, while relatively little work has examined how subjective human psychological states can be mapped to quantitative modulation of these controllers \citep{lee2014neural,kim2019task}. We introduce the psychological state interference function to capture the dynamic influence of different psychological states on goal-directed and habitual control. The purpose of this mechanism is to increase the state dependence and behavioral dynamics of human-like agents, thereby improving the believability and human-likeness of their generated behavior. 
Existing controlled human studies generally show that adverse stress, self-control depletion, or cognitive load is associated with a marked shift in behavioral control, characterized by reduced goal-directed control and stronger habitual responding. Motivated by this empirical trend, we design psychological-state interference, once activated, to induce a behaviorally meaningful shift toward habitual control.
Precisely modeling the mechanisms through which individual psychological factors affect behavioral control is beyond the scope of this work. Instead, inspired by the overall directional trends reported in controlled human studies, we introduce a heuristic mechanism that provides a coarse-grained approximation of these state-dependent shifts in behavioral control. 

At each time step $t$, the agent maintain s a psychological state $s_t^{\mathrm{psy}} = [s_t^{\mathrm{stress}}, s_t^{\mathrm{depletion}}, s_t^{\mathrm{load}}]$, consisting of stress, self-control depletion, and cognitive load. Each factor is represented by an LLM-assigned subjective score on an 11-point Likert-style scale from $0$ to $10$, based on the agent's current state and recent interactions. This representation provides a coarse-grained semantic estimate of the agent's momentary psychological state rather than a standardized measurement from a specific psychological questionnaire. 
We use a score of $3$ as a reference for a typical psychological state and a score of $7$ as an adverse state reference, whose intensity is approximately comparable to the manipulated conditions in the behavioral studies considered below. These values serve as semantic reference points for the state-dependent gate and do not impose an exact baseline-to-adverse control ratio. Scores below $3$ indicate weaker interference, whereas scores above $7$ indicate increasingly stronger psychological interference.

We consider stress, self-control depletion, and cognitive load because they are closely related to behavioral-control allocation. First, these factors have been extensively studied in psychology and cognitive science in relation to shifts between flexible, deliberative goal-directed control and more automatic or habitual responding. Second, they provide suitable empirical references for parameterization. Controlled studies manipulate these factors individually and compare behavioral changes between baseline and elevated conditions, while related studies provide subjective measures for interpreting psychological-state intensity and behavioral or computational measures that reflect changes in goal-directed and habitual control. This allows us to establish an empirical correspondence between changes in individual psychological factors and changes in behavioral-control allocation. For some factors, subjective-state measurements and behavioral-control effects are obtained from different but related studies; we therefore use them separately for state-semantic interpretation and control-effect scaling. Because these factors are examined under heterogeneous experimental paradigms, the reported behavioral measures and absolute control ratios cannot be directly transferred across studies. For each factor $k\in\{\mathrm{stress},\mathrm{depletion},\mathrm{load}\}$, we instead estimate the relative suppression of goal-directed control within each study as $\beta_k=1-G_k^{\mathrm{high}}/G_k^{\mathrm{base}}$, where $G_k^{\mathrm{base}}$ and $G_k^{\mathrm{high}}$ denote empirical proxies for goal-directed control under the baseline and elevated conditions, respectively. 

\paragraph{Stress.} Schwabe and Wolf exposed participants to either a Socially Evaluated Cold Pressor Test or a control condition before an instrumental learning task with extinction blocks \citep{schwabe2011stress}. Subjective stressfulness increased from $3.8$ to $47.2$ on the original $0$--$100$ scale. Stress did not impair overall instrumental learning, but stressed participants persisted more strongly in previously rewarded actions during extinction, indicating reduced sensitivity to changes in action--outcome contingencies. Based on this behavioral persistence, we approximate the Habit:Goal balance as $38{:}62 \rightarrow 66{:}34$, which gives $\beta_{\mathrm{stress}}=1-0.34/0.62\approx0.45$. These Habit:Goal ratios are behavioral proxies rather than controller weights directly estimated in the original study. 

\paragraph{Depletion.} Neal et al.\ studied how reduced self-control affects the expression of strong habits \citep{neal2013people}. In Study~4 ($N=134$), participants completed either a depletion manipulation or a control writing task. On $1$--$7$ subjective scales, ratings of feeling drained increased from $1.99$ to $2.58$, while mental exhaustion increased from $1.94$ to $2.64$. Depletion also increased the probability of choosing strongly habitual options by approximately $31$--$32$ percentage points, while weak habits changed by less than one percentage point. We therefore use the approximate behavioral proxy $16{:}84 \rightarrow 48{:}52$, yielding $\beta_{\mathrm{depletion}}=1-0.52/0.84\approx0.38$. 

\paragraph{Cognitive load.} Otto et al.\ manipulated working-memory availability by adding a concurrent numerical Stroop task to a two-stage reinforcement-learning task \citep{otto2013curse}. Their computational model explicitly mixed model-based and model-free action values using a fitted mixing weight. The model-based contribution was approximately $0.45$ when cognitive resources were relatively available and decreased to approximately $0.27$ under concurrent or recent working-memory load. Using model-based and model-free control as proxies for goal-directed and habit-like control, respectively, gives $55{:}45 \rightarrow 73{:}27$, and therefore $\beta_{\mathrm{load}}=1-0.27/0.45\approx0.40$. We note that model-free reinforcement learning is not strictly equivalent to habitual control and use it only as a computational proxy. Otto et al.\ did not report subjective cognitive-load ratings. We therefore additionally use the subjective workload measurements of Maramotti et al.\ as a semantic reference for this dimension \citep{maramotti2026effort}. In their Stroop experiment, NASA-TLX Mental Demand increased from $5.15$ in the RELAX condition to $6.57$ in the EXERT condition on a $0$--$10$ scale. This result is used only to inform the semantic interpretation of subjective cognitive load, while the controller shift is obtained from Otto et al. The resulting empirical effect-size references are summarized below:
\begin{table}[t] 
\centering 
\caption{Empirical effect-size references for parameterizing the psychological state gate. Habit:Goal ratios are behavioral or computational proxies.} \label{tab:internal_state_gate} \begin{tabular}{lcc} \toprule Factor & Habit:Goal (Normal $\rightarrow$ Elevated) & $\beta_k$ \\ \midrule Stress & $38{:}62 \rightarrow 66{:}34$ & $0.45$ \\ Depletion & $16{:}84 \rightarrow 48{:}52$ & $0.38$ \\ Cognitive load & $55{:}45 \rightarrow 73{:}27$ & $0.40$ \\ \bottomrule \end{tabular} \end{table} 

\paragraph{State-dependent gate activation.} Psychological factors do not deterministically affect behavioral control at every time step. Instead, their quantified scores control the probability that the corresponding interference becomes behaviorally active. 
For each factor $k$, we define a state-dependent activation probability $p_k^{\mathrm{on}}(s)=\sigma(0.55(s-6))$, where $\sigma(\cdot)$ is the sigmoid function. This gives a small but non-zero activation probability at low state levels, e.g., $p^{\mathrm{on}}(0)\approx0.04$ and $p^{\mathrm{on}}(3)\approx0.16$, while the probability increases to $p^{\mathrm{on}}(7)\approx0.63$ and $p^{\mathrm{on}}(10)\approx0.90$ as the psychological-factor score increases. 
At each time step, we sample $z_{t,k}\sim\mathrm{Bernoulli}(p_k^{\mathrm{on}}(s_{t,k}))$, where $z_{t,k}=1$ indicates that the interference associated with factor $k$ becomes behaviorally active at the current decision step. 
Since $\beta_k$ is estimated from average human behavioral effects, we use it as an effect-size reference to set the conditional suppression magnitude so that the expected suppression at $s=7$ matches $\beta_k$.

Specifically, we set $\gamma_k=\beta_k/p_k^{\mathrm{on}}(7)$ and define $f_{t,k}=z_{t,k}\gamma_k$, such that $\mathbb{E}[f_{t,k}\mid s=7]=\beta_k$. With $p^{\mathrm{on}}(7)\approx0.63$, this gives $\gamma_{\mathrm{stress}}\approx0.71$, $\gamma_{\mathrm{depletion}}\approx0.60$, and $\gamma_{\mathrm{load}}\approx0.63$. Existing studies largely manipulate these psychological factors independently and provide limited evidence about their joint effects. We therefore avoid assuming additive interactions and conservatively define the overall interference by the strongest active factor, i.e., $ F(s_t^{\mathrm{psy}}) = \max_{k\in\{\mathrm{stress},\mathrm{depletion},\mathrm{load}\}} f_{t,k}. \label{eq:internal_gate} $ The empirical evidence determines the expected suppression magnitude at the elevated-state reference, while the quantified psychological state score controls the probability that this suppression becomes behaviorally active. Consequently, a higher psychological-factor score produces stronger expected interference on goal-directed control, consistent with the definition of $F(s_t^{\mathrm{psy}})$ in \S \ref{arbiter}.

\section{World model}
\label{world_model}

\begin{figure}
    \centering
    \includegraphics[width=1\linewidth]{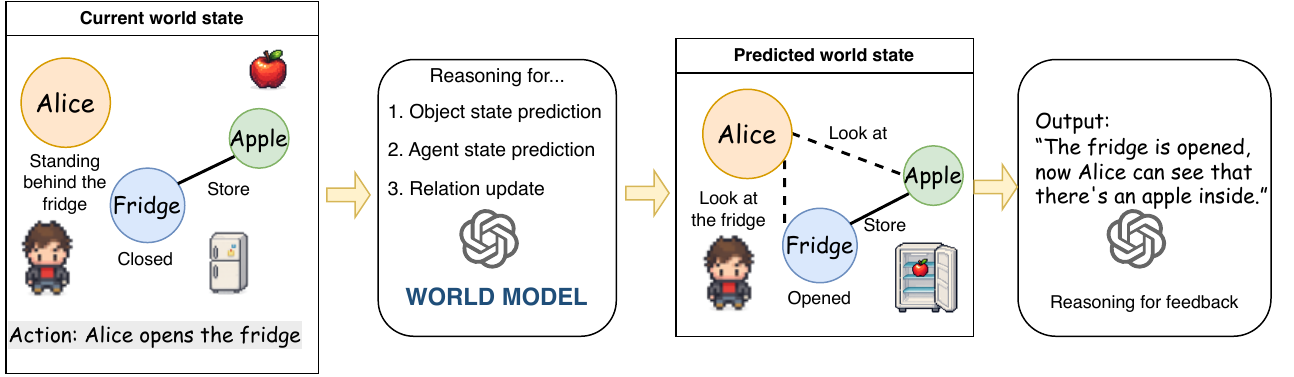}
    \caption{The framework of the context-based world model}
    \label{fig:world_model}
\end{figure}

Inspired by~\citep{zhao2023large,li2026bridging}, we introduce a context-based world model to approximate state transitions for the human-like agent. Specifically, we use a frozen LLM to predict possible next states and action consequences based on the current world state and the agent's action.

As shown in Fig. \ref{fig:world_model}, we represent the world as a graph. Entities in the environment, including the agent and objects, are represented as nodes, while their states are represented as node attributes. Relations and interactions between entities are represented as edges. In this way, each node--edge--node tuple represents a factual relation in the current world state. 

At each state transition, the LLM predicts changes in node attributes and relations between entities. These changes represent how the world may evolve after an action is executed. The predicted updates are then organized into a structured JSON format and applied to the current world representation to obtain the predicted next state.

The context-based world model has several limitations. First, its predictions depend on the domain knowledge of the underlying LLM, and prediction errors may accumulate over long trajectories. Second, the frozen LLM does not dynamically update its world knowledge during interaction. However, our setting mainly focuses on daily-life activities, for which LLMs generally provide sufficient prior knowledge to predict common action consequences. We therefore adopt a simple frozen LLM as the agent's internal model. This design also reflects a property of human internal models, whose predictions are similarly constrained by prior knowledge and experience.


\section{Text to motion based 3D embodied module}
\label{3Dmodule}

\paragraph{3D Module Overview}
The 3D module translates the natural language action description into human motion within an interactive scene. Following the training-free scene-aware Text-to-Motion paradigm~\citep{tstmotion}, the module grounds high-level behaviors in 3D scenes through inference-time planning and spatial constraints without retraining the underlying motion generator. The module comprises a high-level planner, keyframe constraints, motion generation, and 3D execution. Given an action description and the current character avatar and scene state, the high-level planner organizes the intended behavior into an ordered sequence of executable actions.

Based on factors such as action complexity, interaction detail, and spatial location, the planned actions are categorized into two types: actions that require spatial guidance and standard actions that do not require guidance. For those actions requiring spatial guidance, the high-level planner provides sparse keyframe constraints that specify relevant body configurations and interaction relationships. These constraints guide motion generation alongside the action descriptions. Actions that do not require spatial guidance are generated directly from the motion generator.

The generated motions along with associated object interactions are executed in the 3D simulation module. The character avatar and scene states are carried forward across successive actions to support multi-step behavior. The following subsections describe the execution setup and examine the contribution of keyframe constraints through qualitative ablations.

\paragraph{Scene Representation and State Management}
The 3D environment is represented by a structured scene description containing object identities, spatial information, and interaction states. Object identities provide consistent references across task planning and execution, while spatial information describes the locations and geometric properties of scene objects. Together with the character avatar's current position and pose, this representation provides the context for grounding behavioral descriptions in the environment.

The scene information is refreshed at each task initialization. The information is sent to the cognitive module and high-level planner for the planning and action generation process. The module maintains both the character avatar state and the object states throughout execution. The character state includes the current body configuration and ongoing interactions, such as facing direction or holding an object. The object state includes changes resulting from these interactions, such as an updated position or an opened door. Subsequent actions receive the preceding action’s terminal state, allowing them to continue from the established configuration.

During continuous execution, the framework carries forward the accumulated interaction state and incorporates updates from the scene. The 3D simulation module provides scene data and executes the corresponding motions and object interactions, while the planner and cognitive module use this information to prepare subsequent actions.

\paragraph{Semantic Task Decomposition and Execution Planning}

Given an action description, the character avatar state, and the environment states, the high-level planner decomposes the coarse-grained intended behavior into an ordered sequence of executable actions. For each executable action, the planner clarifies how to perform it, identifies relevant interaction objects, and specifies their expected outcomes.

The planner considers spatial relationships and the character avatar's current states when planning the action sequence. Movement actions are included when needed to reach an interaction location, while actions that can be performed from the current position retain that position where appropriate. 
Affordance is also considered when determining whether preparatory actions are needed. For example, pulling out the chair will be planned when there is not enough sitting space under the table. Such adjustments remain part of semantic planning, ensuring that the resulting sequence forms a coherent and spatially feasible course of action.

Execution planning translates the resulting action sequence into inputs for the Low-Level Motion Generator. It aligns action descriptions with applicable keyframe constraints, assigns their timing, and schedules interaction events. Sequences exceeding the motion generator’s per-call duration are divided into consecutive generation batches while preserving the planned action order. Boundary states are carried across batches to support continuity.

\paragraph{Keyframe Constraints for Scene-Grounded Motion}
In the current stage, we use NVIDIA Kimodo \cite{rempe2026kimodo} as the Low-Level Motion Generator.  Although Kimodo provides stable and high-quality action generation capabilities, it lacks the ability to perceive scene information. In order to generate the action that can precisely interact with objects in the scene, we use keyframe constraints to guide Kimodo to generate scene-grounded motions.

Given an intended behavior description and the current environment states, which include the character avatar state, the keyframe generation module produces sparse constraints that guide the action generator in generating scene-matched actions. For example, opening a door requires placing the hand at the handle’s spatial coordinates, while sitting requires positioning the body over the seat in an appropriate posture and orientation. Constraints also specify the avatar’s facing direction relative to a target object. In general, generated keyframes provide spatial reference states for the intended behavior, while the low-level motion generator produces the movement between them.

The resulting keyframes are aligned with the corresponding action descriptions during execution planning and jointly supplied to the motion generator. For continuous tasks, boundary states are propagated across successive actions to support continuity.

\paragraph{Keyframe Constraints Study}
To better understand the contribution of keyframe constraints, we conduct a keyframe ablation study on a set of common daily activities. Each task is evaluated using both the full module and an ablated variant in which the keyframe constraint module is disabled, while the low-level motion generator and action descriptions remain unchanged. In this ablation, basic navigation and positional guidance are retained, including the coordinates of the target object or support surface and the associated body-joint relationships. Only the additional keyframe-level pose and interaction constraints are removed: the ablated variant receives no explicit keyframe poses, temporally aligned contact states, or keyframe-level guidance for character position and orientation. This setting allows us to isolate the contribution of keyframe constraints while retaining the minimum information required for scene-aware motion generation.

From the complete set of generated results, we select four representative cases for qualitative visualization: (1) sitting on a sofa while watching television and subsequently standing up to cheer, (2) sitting at a dining table, (3) performing jumping jacks in a suitable open area in the living room, and (4) walking toward a toilet while holding the abdomen in pain and then sitting on the toilet. These cases cover different forms of spatial guidance, including body orientation, placement relative to scene objects, positioning in an open area, transitions between supported and standing poses, and locomotion accompanied by a specified body gesture. The first and fourth cases are each presented using four snapshots to illustrate multiple stages of the behavior, while the second and third cases are shown as paired comparisons. Together, these examples provide a qualitative comparison of the motions generated with and without keyframe constraints.

\noindent\textbf{Scenario 1: [Sitting on a sofa while watching television and subsequently standing up to cheer].}
\begin{figure}
\begin{subfigure}{0.485\linewidth}
    \centering
    \includegraphics[width=\linewidth]{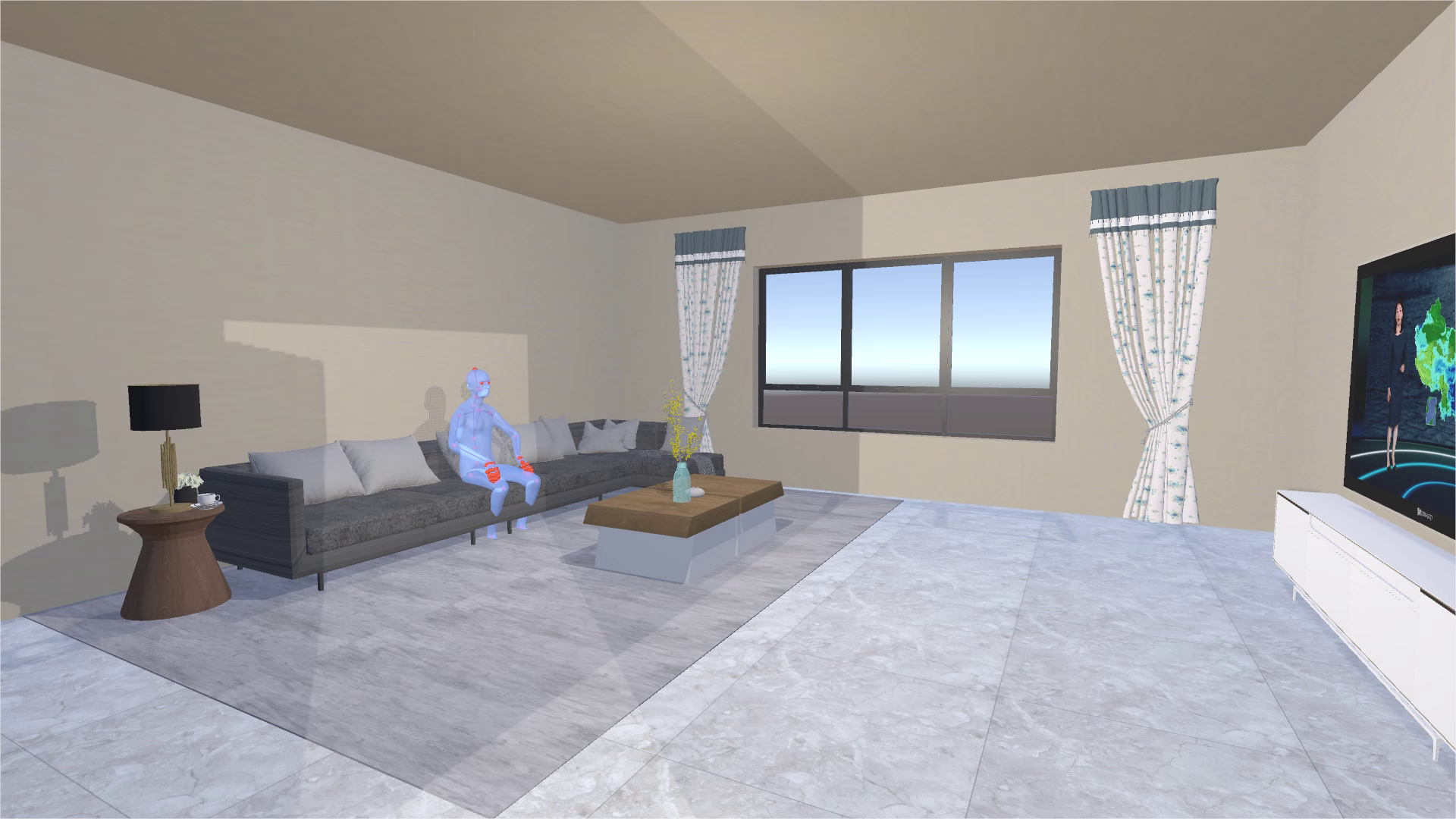}
    \caption{Full Module}
    \label{fig:sofasit}
\end{subfigure}
\hfill
\begin{subfigure}{0.485\linewidth}
    \centering
    \includegraphics[width=\linewidth]{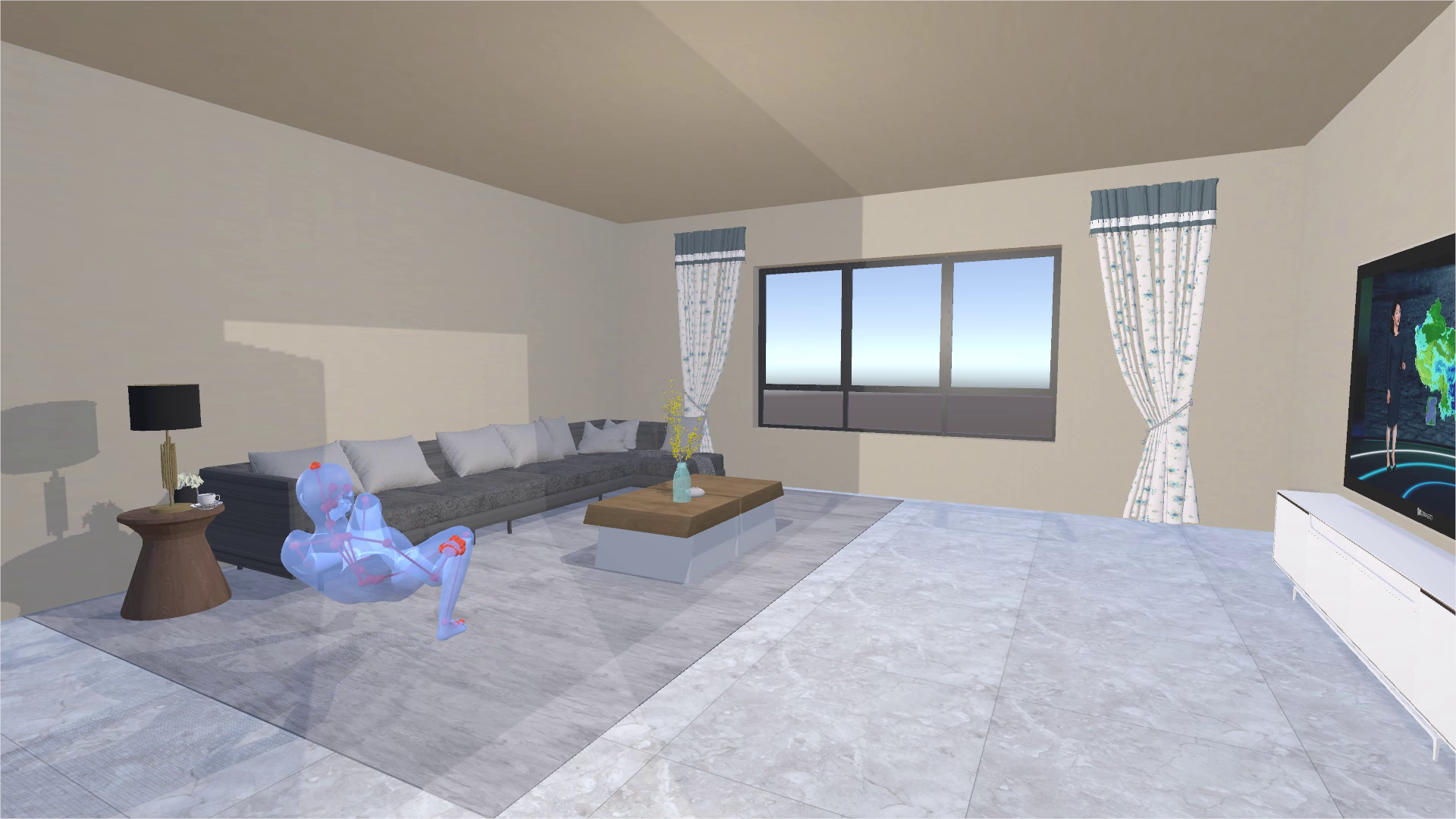}
    \caption{Without keyframe constraints}
    \label{fig:sofasit-ab}
\end{subfigure}
\caption{Ablation task: Sitting on the sofa while watching television}
\label{fig:keyab-sofa}
\end{figure}

\begin{figure}
\begin{subfigure}{0.485\linewidth}
    \centering
    \includegraphics[width=\linewidth]{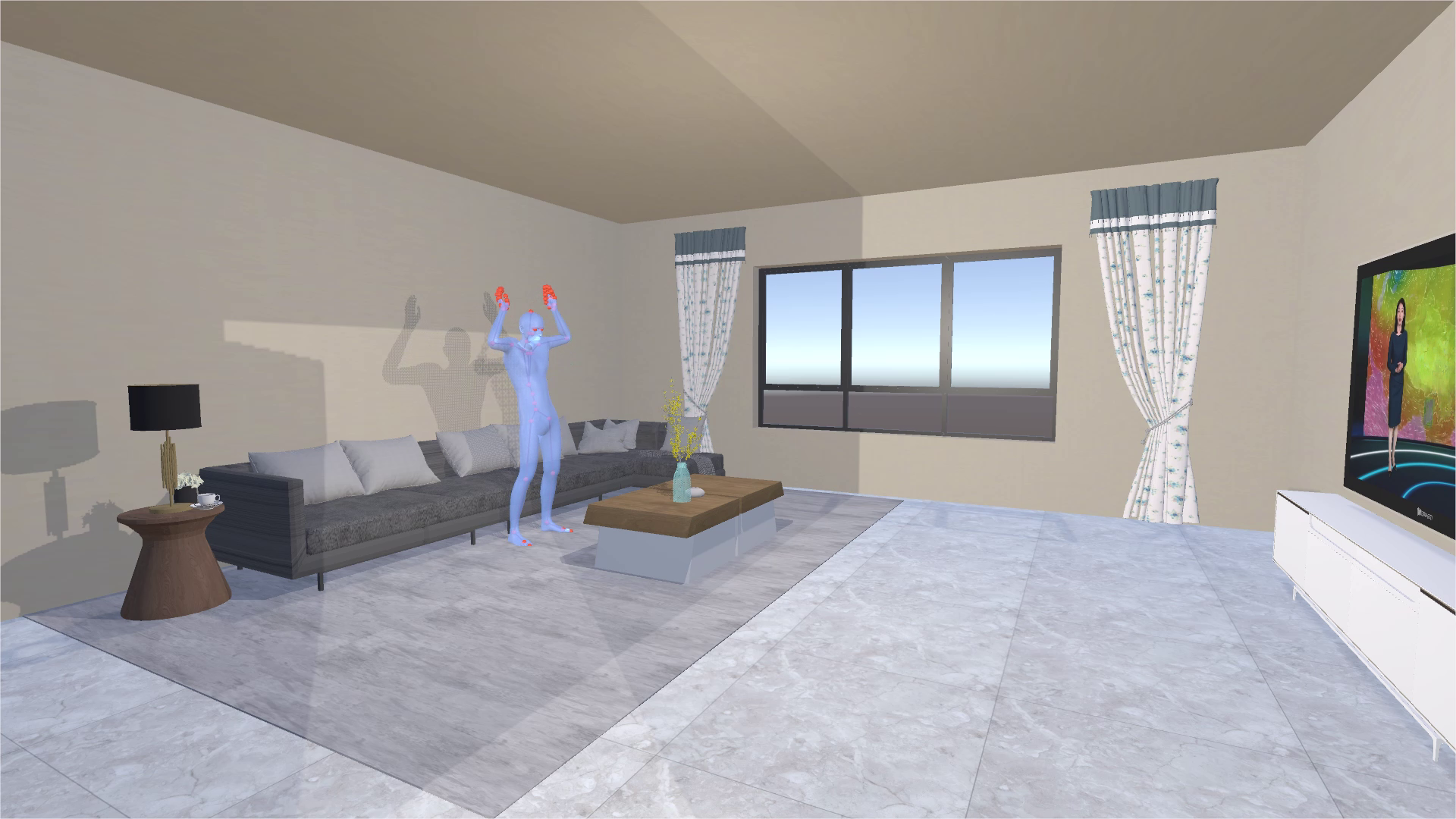}
    \caption{Full Module}
    \label{fig:cheer}
\end{subfigure}
\hfill
\begin{subfigure}{0.485\linewidth}
    \centering
    \includegraphics[width=\linewidth]{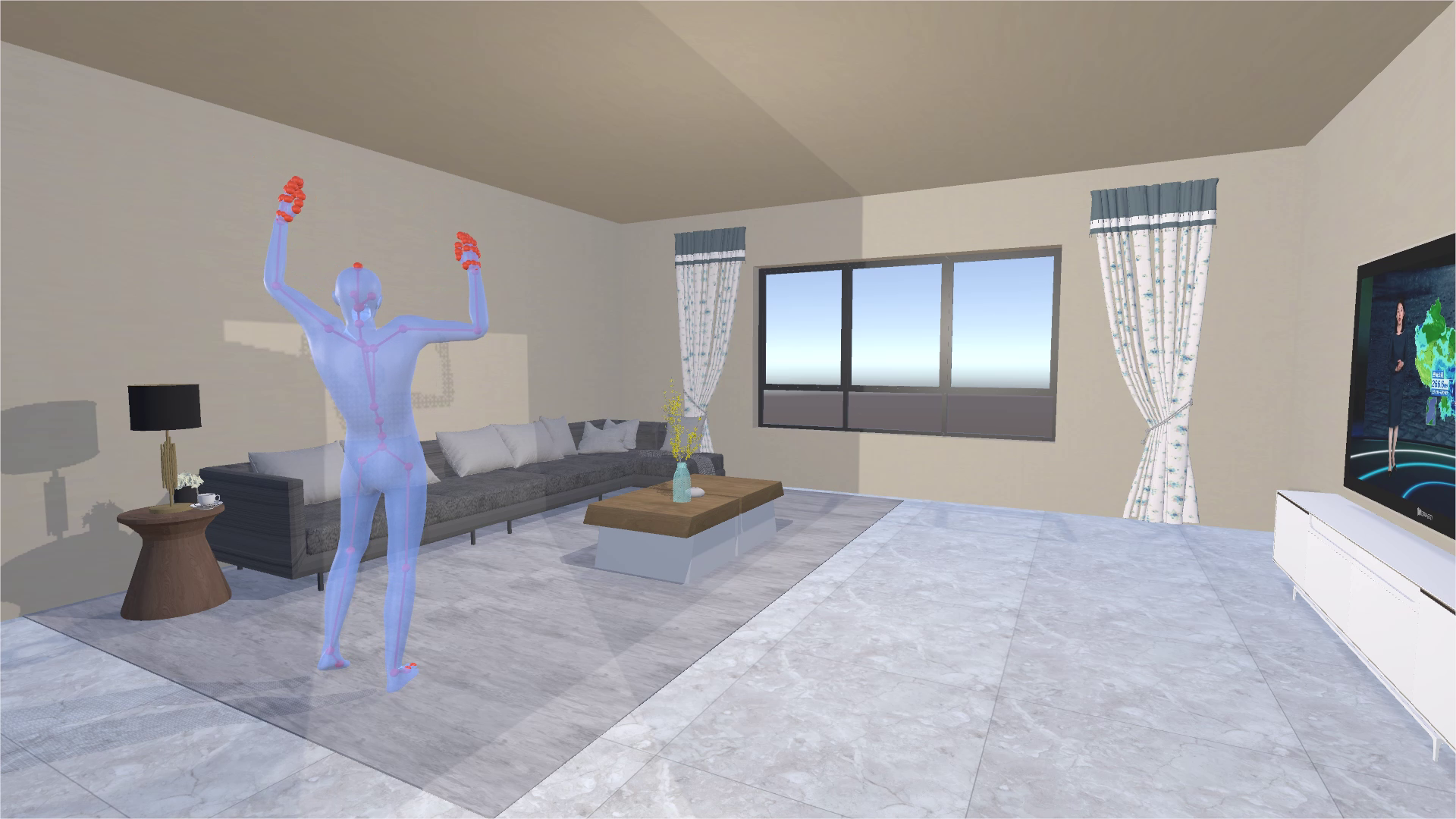}
    \caption{Without keyframe constraints}
    \label{fig:cheer-ab}
\end{subfigure}
\caption{Ablation task: standing up to cheer}
\label{fig:keyab-cheer}
\end{figure}

As shown in Figs.~\ref{fig:keyab-sofa} and~\ref{fig:keyab-cheer}, in the ablated setting, navigation brings the character to the vicinity of the sofa, but the character remains too far away to establish physical support from the seat. Without keyframe guidance describing the spatial relationship between the body and the sofa geometry, the generator produces an implausible sitting–reclining pose: the feet remain on the floor, while the upper body assumes a reclined posture without being supported by the sofa. The character’s gaze is also not aligned with the television. By comparison, the full module places the character on the sofa with an appropriate supported pose and viewing direction. The subsequent self-interaction component of cheering is successfully generated in both settings.

\noindent\textbf{Scenario 2: [Sitting at a dining table].}
\begin{figure}
\begin{subfigure}{0.485\linewidth}
    \centering
    \includegraphics[width=\linewidth]{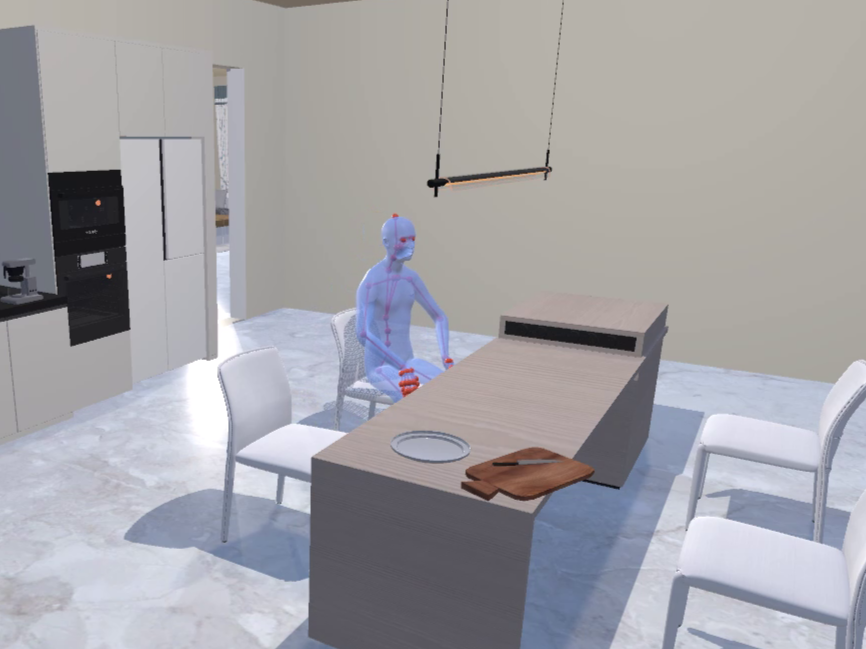}
    \caption{Full Module}
    \label{fig:sitTable}
\end{subfigure}
\hfill
\begin{subfigure}{0.485\linewidth}
    \centering
    \includegraphics[width=\linewidth]{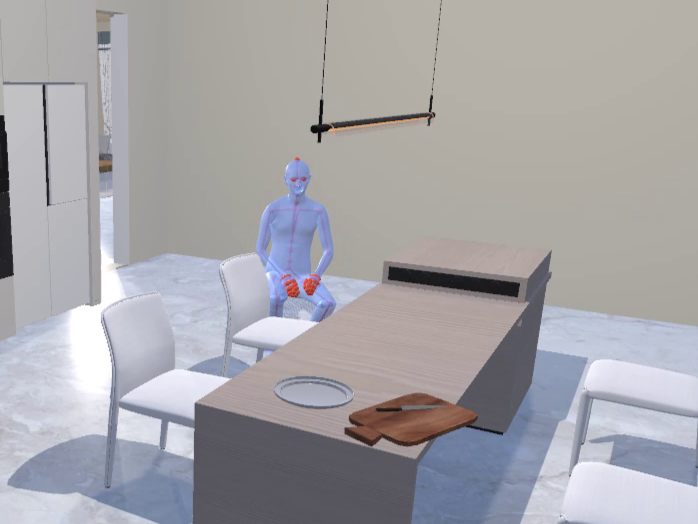}
    \caption{Without keyframe constraints}
    \label{fig:sitTable(ab)}
\end{subfigure}
\caption{Ablation task: Sitting in front of the kitchen table}
\label{fig:keyab-sitTable}
\end{figure}

 In the second comparison, as can be seen from Fig. \ref{fig:keyab-sitTable}, the navigation component successfully brings the character into the vicinity of the target in both settings. However, target proximity alone is insufficient to establish a valid sitting configuration. Without keyframe constraints, the ablated model lacks the scene-level geometric guidance needed to interpret the chair’s supporting region and functional orientation. Even when the coordinates of the sittable surface are provided, the character places the pelvis beside the supporting region and adopts a sitting direction unrelated to the orientation of the chair. This results in a partially unsupported and spatially inconsistent sitting pose. In contrast, the full module aligns both the body position and orientation with the seat, producing a properly supported sitting pose that is consistent with the chair and its surrounding scene.

\noindent\textbf{Scenario 3: [Performing jumping jacks in a suitable open area in the living room].}
\begin{figure}
\begin{subfigure}{0.485\linewidth}
    \centering
    \includegraphics[width=\linewidth]{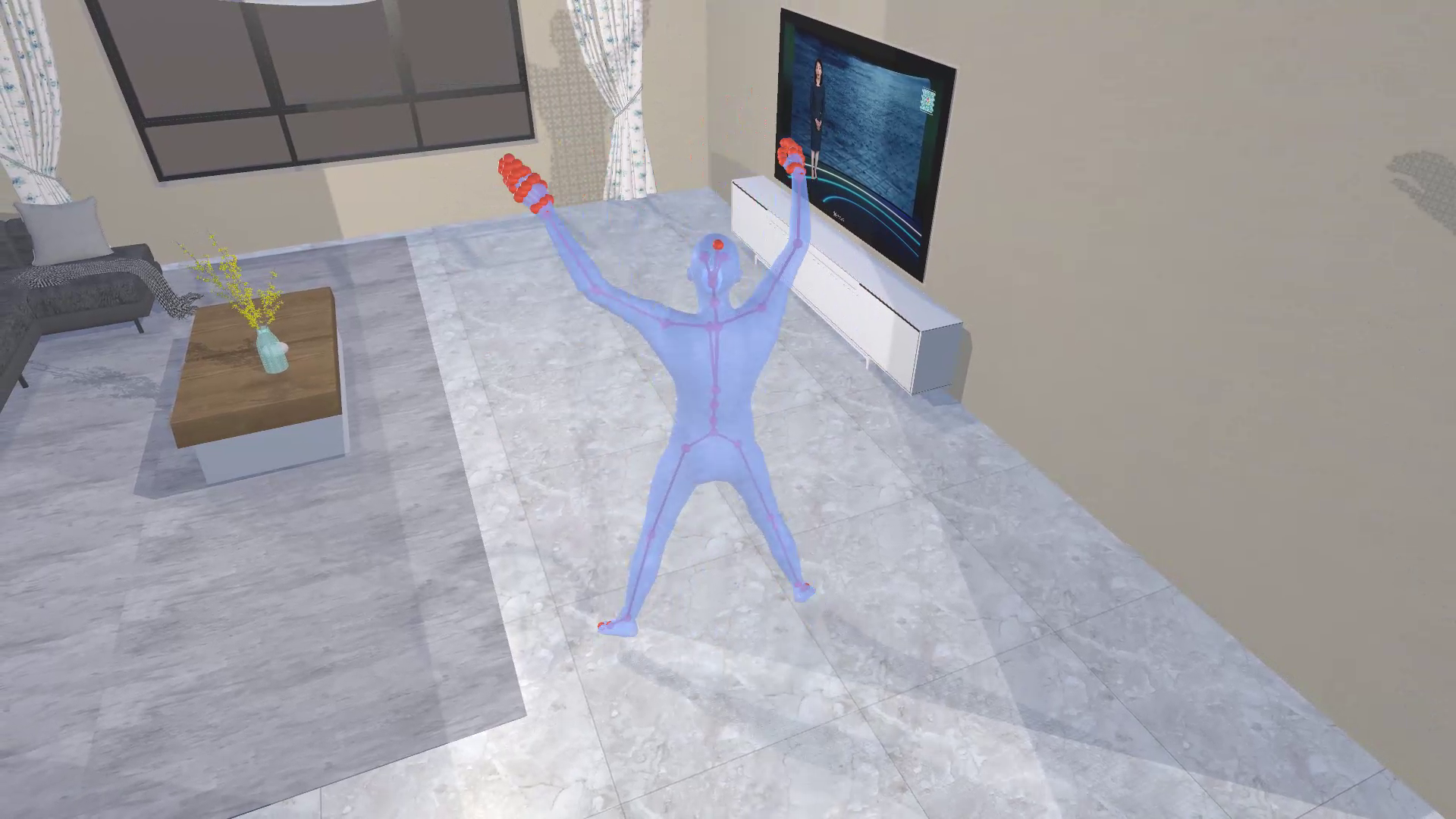}
    \caption{Full Module}
    \label{fig:jumpingJacks}
\end{subfigure}
\hfill
\begin{subfigure}{0.485\linewidth}
    \centering
    \includegraphics[width=\linewidth]{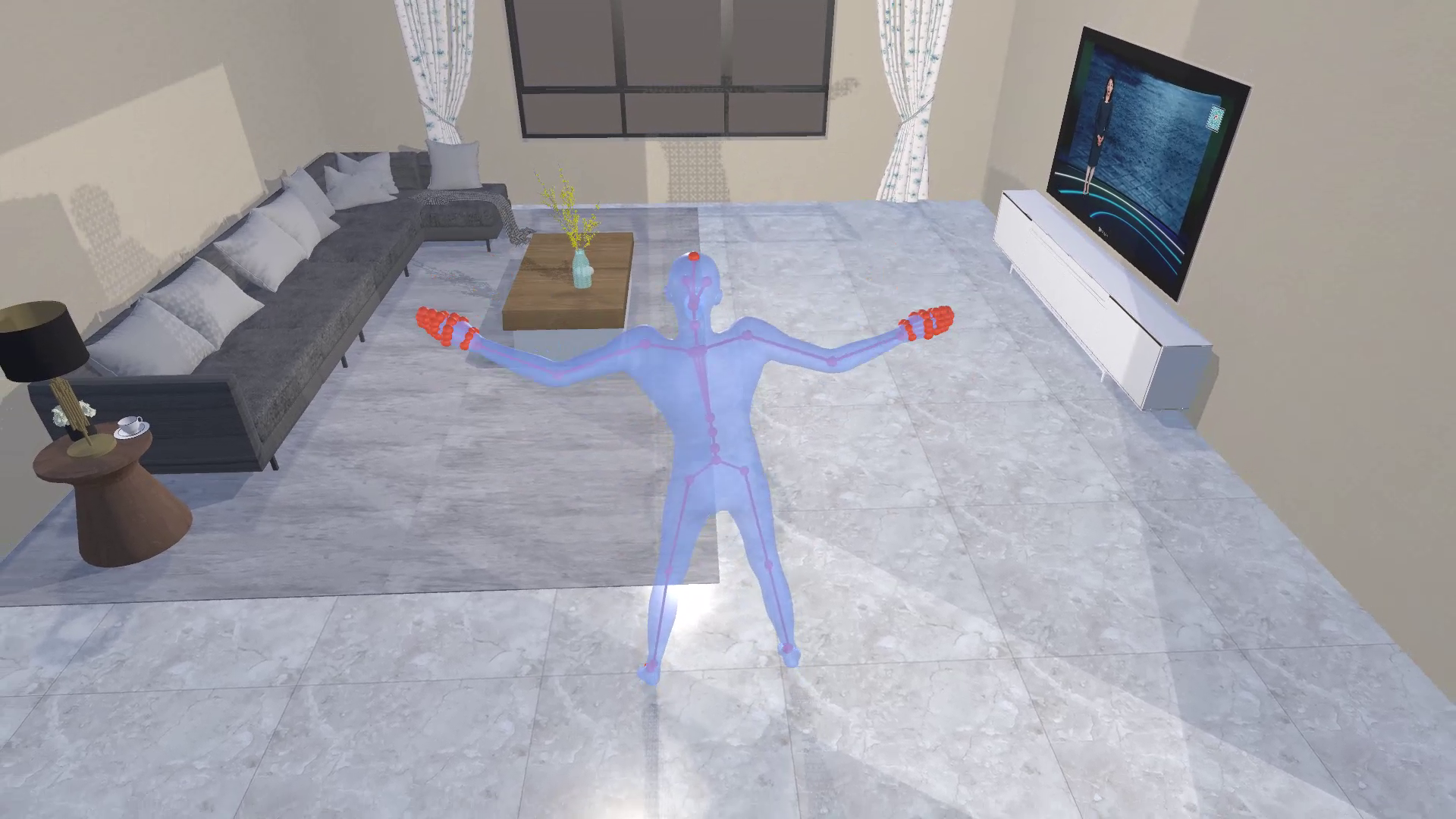}
    \caption{Without keyframe constraints}
    \label{fig:jumpingJacks(ab)}
\end{subfigure}
\caption{Ablation task: Performing jumping jacks in a suitable open area in the living room}
\label{fig:keyab-jumpingJacks}
\end{figure}

As shown in Fig. \ref{fig:keyab-jumpingJacks}, both settings correctly interpret the non-object-specific navigation target, identify a suitable open area in the scene, and successfully navigate the character to the selected location. After reaching the open area, both the full module and the ablated variant successfully generate the requested jumping-jack motion.

\noindent\textbf{Scenario 4: [Walking toward a toilet while holding the abdomen in pain and then sitting on the toilet].}

\begin{figure}[t]
\begin{subfigure}{0.485\linewidth}
    \centering
    \includegraphics[width=\linewidth]{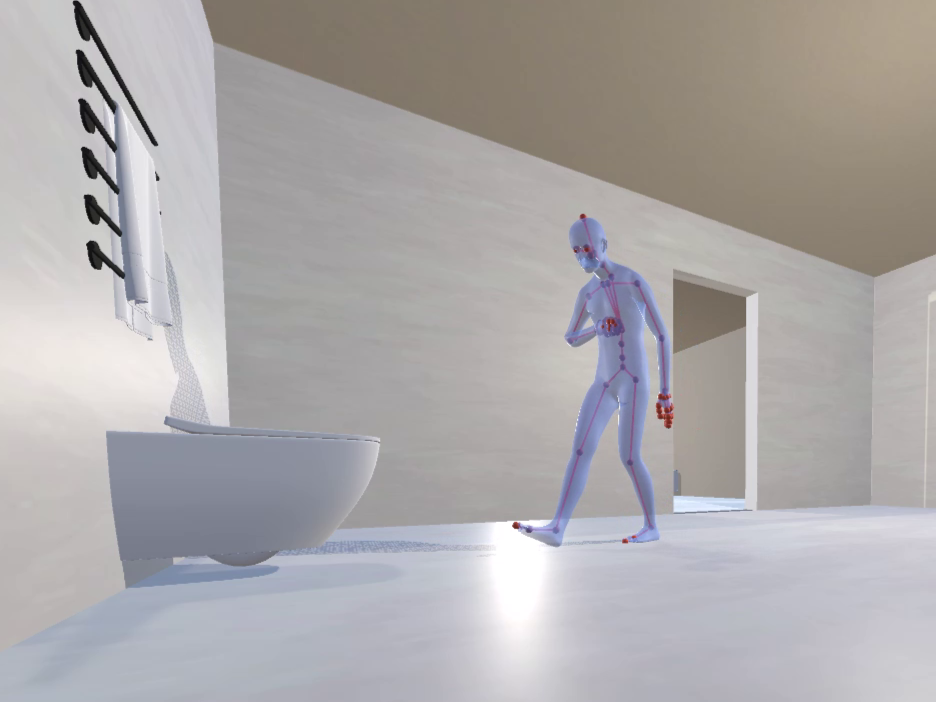}
    \caption{Full Module}
    \label{fig:PainWalk}
\end{subfigure}
\hfill
\begin{subfigure}{0.485\linewidth}
    \centering
    \includegraphics[width=\linewidth]{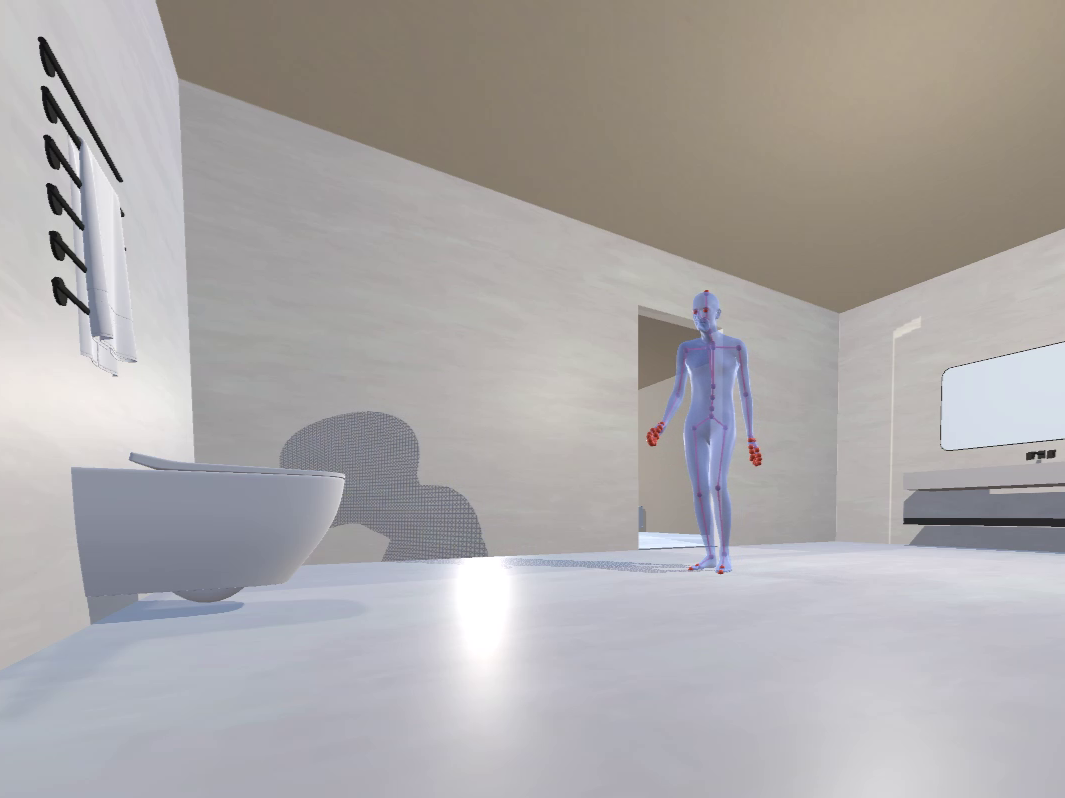}
    \caption{Without keyframe constraints}
    \label{fig:PainWalk(ab)}
\end{subfigure}
\caption{Ablation task: Walking toward a toilet while holding the abdomen in pain}
\label{fig:keyab-PainWalk}
\end{figure}

\begin{figure}[t]
\begin{subfigure}{0.485\linewidth}
    \centering
    \includegraphics[width=\linewidth]{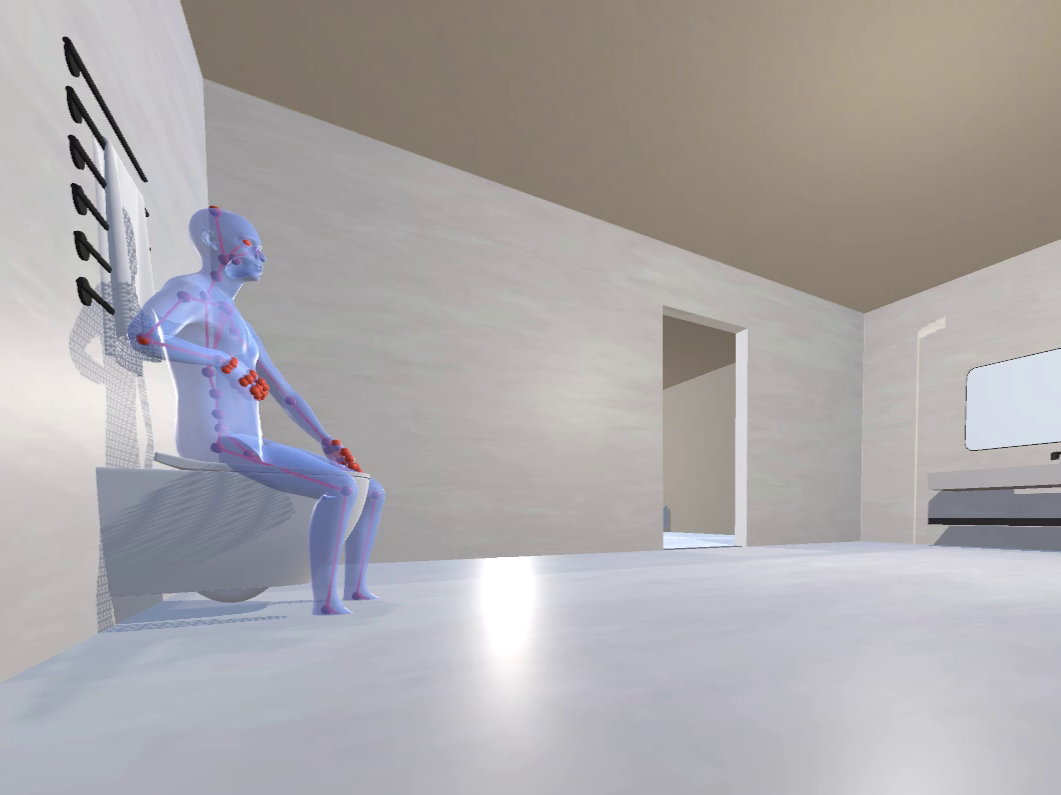}
    \caption{Full Module}
    \label{fig:toiletSit}
\end{subfigure}
\hfill
\begin{subfigure}{0.485\linewidth}
    \centering
    \includegraphics[width=\linewidth]{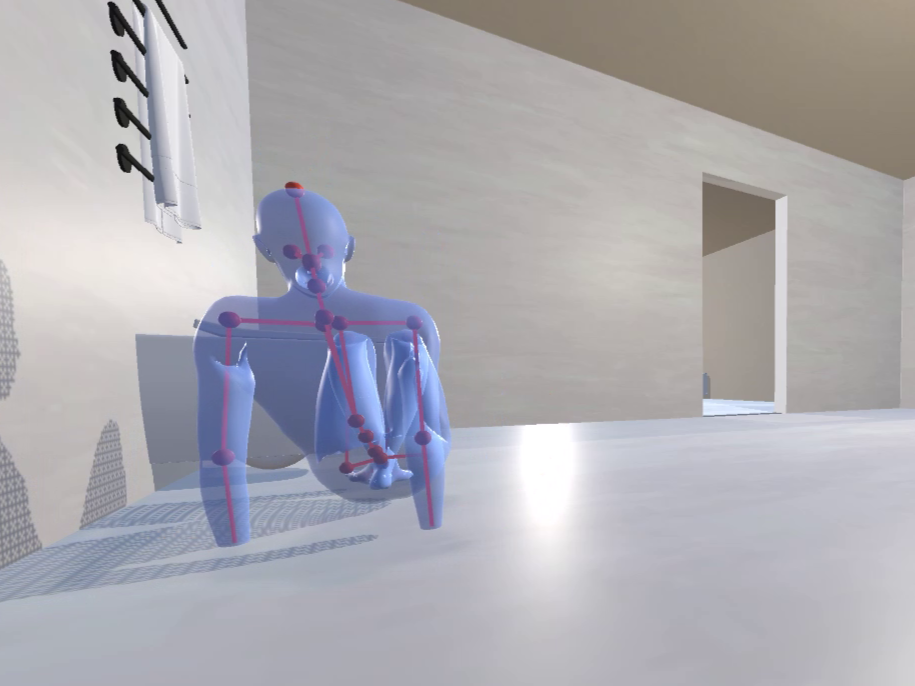}
    \caption{Without keyframe constraints}
    \label{fig:toiletSit-ab}
\end{subfigure}
\caption{Ablation task: Sitting on the toilet}
\label{fig:keyab-toiletSit}
\end{figure}

The fourth case, as shown in Figs.~\ref{fig:keyab-PainWalk} and~\ref{fig:keyab-toiletSit}, walking toward a toilet while holding the abdomen in pain, differs from the preceding examples because the specified body configuration must be maintained throughout locomotion. Without keyframe guidance, Kimodo produces a plausible but generic walking pattern, in which the abdomen-holding component is no longer visibly expressed. The full module supplements the action description with spatial pose guidance, enabling the requested body configuration to remain identifiable during walking. In the subsequent sitting stage, the ablated variant fails to establish the required spatial relationship with the toilet and lowers the character directly onto the floor. In contrast, the full module positions the character relative to the toilet and completes the intended sitting action.

\paragraph{Keyframe Constraints Study Conclusion}
Across the four scenarios, disabling keyframe constraints does not noticeably affect the navigation component: the character can still identify the intended region and move to the vicinity of the target. However, the ablated module largely loses the ability to establish reliable interactions with external objects, particularly when the action requires precise spatial relationships among the character, the target object, and the surrounding scene. Self-interaction actions are generally less affected because they depend less on external geometry. Nevertheless, when an action description does not reliably elicit the requested motion pattern from the low-level generator, the ablated module may fall back to a more generic action. The full module provides more consistent scene grounding and better preserves the intended action characteristics in these cases. Its motion quality, however, remains partially dependent on the quality of the generated keyframes; strongly constrained poses, such as some sitting configurations, may occasionally appear slightly less natural.

\end{document}